%% file: main.tex
\documentclass[11pt]{article}

\usepackage[final]{acl}

\usepackage{times}
\usepackage{latexsym}
\usepackage[T1]{fontenc}
\usepackage[utf8]{inputenc}
\usepackage{microtype}
\usepackage{inconsolata}
\usepackage{graphicx}
\usepackage{url}
\usepackage{amsmath}
\usepackage{multirow}
\usepackage{booktabs}
\usepackage{enumitem}
\usepackage{amsfonts}
\usepackage{adjustbox}
\usepackage{tcolorbox}

\newcommand{\uuline}[1]{%
  \underline{\underline{#1}}%
}

\newcommand{\uline}[1]{%
  \underline{#1}%
}

\title{SoftGene: Protein Language Model-Enhanced Soft Prompting\\ for Interpretable Gene Set Annotation}

\author{
Drew Ross$^{\dagger}$,
Arya Hadizadeh Moghaddam$^{\dagger}$,
Dongjie Wang$^{\dagger}$,\\
\textbf{Xiaoyu Zhang$^{\ddagger}$,
Zijun Yao$^{\dagger}$\thanks{Corresponding author.}}
\\[6pt]
  $^{\dagger}$Electrical Engineering and Computer Science, University of Kansas, USA\\
\texttt{\{drewross, a.hadizadehm, wangdongjie, zyao\}@ku.edu}\\
[3pt]
$^{\ddagger}$Cell Biology and Physiology, University of Kansas Medical Center, USA \\
\texttt{xzhang8@kumc.edu}
}

\begin{document}
\maketitle
\input{sections/abstract}
\input{sections/introduction}
\input{sections/methodology}

\input{sections/experiments}
\input{sections/relatedworks}
\input{sections/conclusion}
\section{Ethical Considerations}
SoftGene should not be used as a substitute for expert biological judgment or experimental validation, particularly in high-stakes biomedical or clinical settings. We encourage users to treat generated annotations as hypothesis-generating tools rather than definitive functional claims, and future work should explore mechanisms for uncertainty estimation, expert-in-the-loop evaluation, and bias-aware benchmarking in gene set interpretation.
\section{Limitations}
\noindent\textbf{Dataset Coverage}: SoftGene is trained and evaluated on curated resources biased toward well-studied genes and biological processes, which may limit generalization to poorly annotated or emerging functions. Moreover, the contribution of protein-sequence embeddings varies across GO domains. Future work should evaluate on less curated gene sets and further characterize when sequence information provides the greatest benefit.

\noindent\textbf{Evaluation Scope}: Although we supplement lexical metrics with biomedical semantic and ontology-aware measures, these evaluations remain reference-based and do not fully capture biological correctness or experimental validity. Future work should incorporate expert evaluation and downstream biological validation to more rigorously assess generated annotations.

\section{Acknowledgments}
This work is dedicated to the memory of Drew Ross and to his family. Drew’s brilliance, warmth, and dedication drove this project from its inception, and he is deeply missed by all of us. 

This work was supported by the National Science Foundation under Grant No. OIA-2531881.
\begin{NoHyper}
\bibliography{references}
\end{NoHyper}
\input{sections/appendices}
\end{document}

%% file: sections/abstract.tex
\begin{abstract}
Gene set analysis is a cornerstone of functional genomics, yet it remains labor-intensive and heavily dependent on manual curation and expert biological interpretation.
While Large Language Models (LLMs) have emerged as powerful tools for genomic reasoning and annotation, most existing approaches rely on symbolic gene names and fail to capture domain-specific biological structure, particularly protein sequence information that governs molecular activity, interactions, and downstream gene function.
In this work, we propose \textbf{SoftGene}, a novel framework for LLM-based gene set annotation that leverages the hierarchical structure of gene sets.
First, we use a hierarchical attention-based encoder built on ESM, a protein language model, to represent each gene set using protein-level amino acid sequence information.
Second, we construct a hybrid prompting scheme that combines soft prompts derived from gene set embeddings with hard prompts containing auxiliary context generated by an LLM, and feed the resulting prompt into a local LLM for annotation.
We evaluate our framework on two benchmark datasets: Gene Ontology (GO) and the Molecular Signatures Database (MSigDB).
Our results show that integrating protein-sequence representations with textual context improves gene set annotation overall, while per-domain analyses reveal that the contribution of protein embeddings varies across biological domains.
\end{abstract}

%% file: sections/introduction.tex
\section{Introduction}
\begin{figure*}[t]
    \centering
    \includegraphics[width=0.90\textwidth]{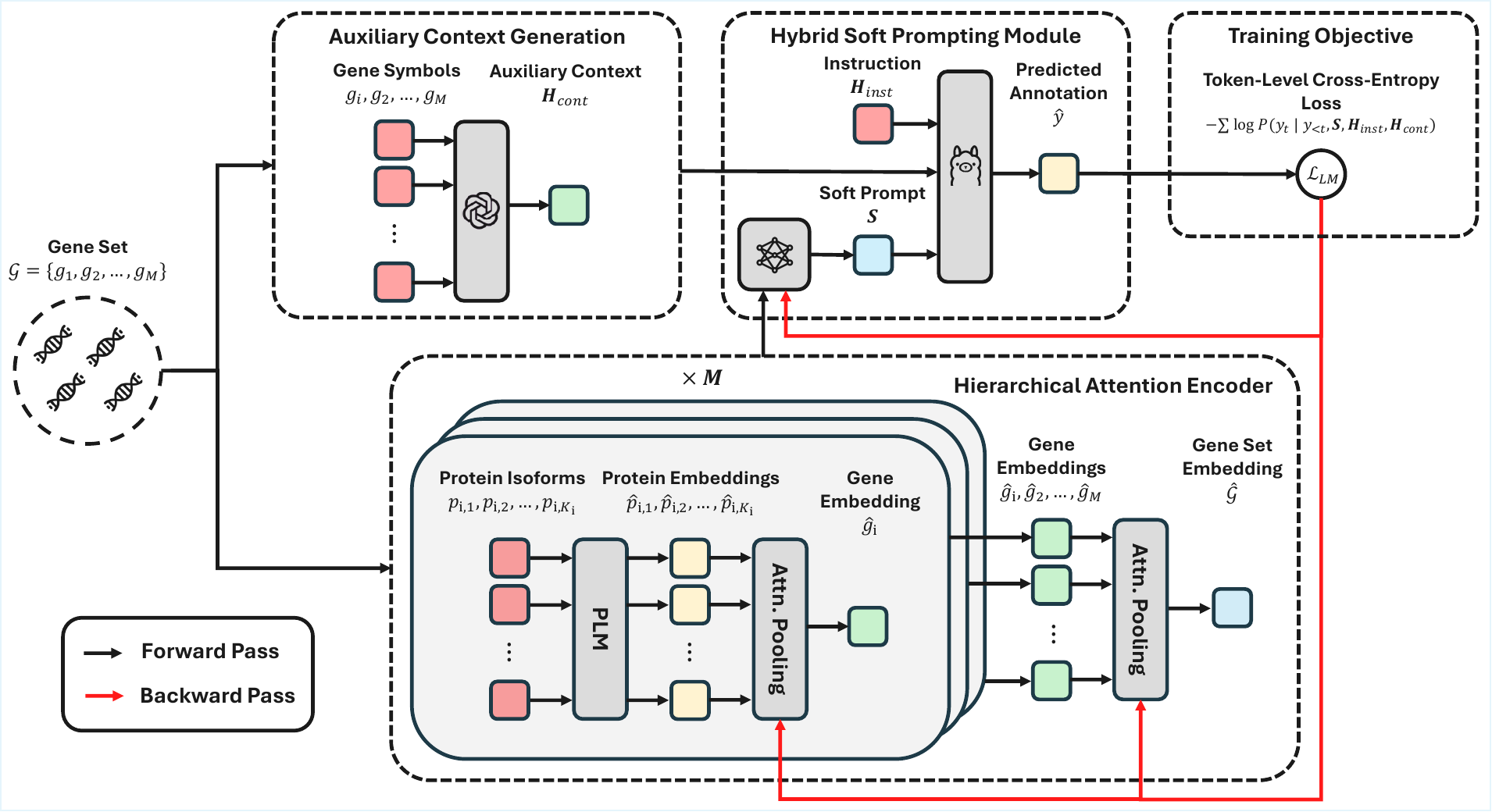}
    \vspace{-0.2cm}
    \caption{High-level overview of the model architecture. The proposed approach combines a sequence-level pathway that encodes amino acid information into soft prompt tokens with a textual pathway that provides high-level functional context from a large language model, enabling the generation of descriptive gene set annotations optimized via supervision against ground-truth labels.}
    \label{fig:overview}
    \vspace{-0.3cm}
\end{figure*}
Gene sets represent cohorts of functionally related genes that collectively drive specific biological pathways and cellular processes \cite{ashburner2000gene}. Generating gene set annotations, which are concise descriptions that delineate shared functional behaviors and metabolic roles, is a fundamental task in modern genomics \cite{ji2026annotating}. As sequencing continues to reveal novel gene associations, the ability to rapidly and accurately annotate these sets is essential for translating raw data into biological insights \cite{salzberg2019next}.

Traditionally, gene set annotation has relied on expert curation or statistical enrichment methods, such as Over-Representation Analysis (ORA) \cite{beissbarth2004gostat} and Gene Set Enrichment Analysis (GSEA) \cite{subramanian2005gene}. However, manual annotation is a time-consuming process that cannot scale with the current pace of data generation \cite{baumgartner2007manual}. Furthermore, existing automated methods are inherently limited by their dependence on curated databases, which are often incomplete or biased toward well-studied genes and biological processes \cite{stoeger2018large, haynes2018gene}.

The recent emergence of Large Language Models (LLMs) has introduced a promising paradigm for a wide range of biological and biomedical tasks \cite{luo2022biogpt, jin2024genegpt, ferruz2022protgpt2} including automated gene annotation \cite{wang2024geneagentselfverificationlanguageagent, joachimiak2024genesetsummarizationusing}.
However, current LLM-based approaches often overlook the domain-specific biological structure of constituent proteins. Protein-sequence representations can provide evolutionary and structural signals such as subcellular localization \cite{thumuluri2022deeploc}, interaction-relevant interfaces \cite{sun2025esm2_amp}, 
and structural properties \cite{simon2025interplm} that may complement symbolic gene information \cite{anfinsen1973principles}. The utility of these sequence-derived signals, however, can vary across biological domains.
Moreover, hard-prompting strategies typically treat genes as semantic tokens, and LLMs remain prone to hallucination \cite{kalai2025language}, which reduces their reliability for this task.
Lastly, although many LLM approaches provide narrative explanations through generative reasoning \cite{hu2025evaluation}, they still lack a concrete decision path showing how specific genes or protein domains contribute to the predicted annotation.

In this paper, we address these limitations by presenting a soft-prompting framework for interpretable gene set annotation. An overview of the framework is shown in Figure \ref{fig:overview}. Our main contributions are as follows:

\begin{itemize}[topsep=0pt,leftmargin=*,nosep]
    \item  We incorporate the embeddings from a pretrained protein language model (PLM) and introduce a hierarchical attention encoder that learns to weigh the relative importance of biological features and aggregate them into a comprehensive representation of gene sets.
    \item  We propose a conditioning scheme that bridges biological sequence data and natural language, enabling an LLM to generate annotations informed by both text-based context and protein-sequence-derived representations.
    \item  We provide concrete interpretability by leveraging the hierarchical attention mechanism to quantify the relative importance of biological features contributing to each annotation. This yields explicit, feature-level explanations grounded in the input data, rather than relying solely on implicit reasoning within the language model.
    \item  We evaluate our framework on two widely used benchmark datasets: Gene Ontology (GO) and MSigDB. Results demonstrate superior performance over existing baselines across metrics.
\end{itemize}

%% file: sections/methodology.tex
\section{Methodology}
In this section, we formalize the gene set annotation task and introduce the associated notation. We then provide an overview of the proposed model architecture, followed by a detailed description of each component of the method.
\subsection{Problem Formulation}
A gene set is a collection of genes that are grouped together based on shared biological function, participation in a common pathway, or co-regulation.
Gene sets are widely used in genomics to represent higher-level biological processes and functional modules.

Let $\{\mathcal{G}_1, \dots, \mathcal{G}_N\}$ denote a collection of gene sets. Each gene set $\mathcal{G}$ consists of a variable number of member genes $\mathcal{G} = \{g_1, \dots, g_M\}$ where $M$ is the length of the set. The ordering of genes within a set is assumed to be irrelevant. Each gene $g_i$ encodes one or more proteins $\{p_{i,1}, \dots, p_{i,K_i}\}$ where $K_i$ represents the number of protein isoforms associated with $g_i$; $K_i$ may vary across genes, reflecting multiple isoforms or splice products. Each protein $p_{i,j}$ is represented as a sequence of amino acids
\begin{equation}
p_{i,j} = (a_1, a_2, \dots, a_{L_{i,j}}),
\end{equation}
where $L_{i,j}$ denotes the length of the protein and each $a_\ell \in \mathcal{A}$ is an amino acid from the standard set of 20 residues, $\mathcal{A} = \{\text{A}, \text{C}, \dots, \text{Y}\}$. 

For each gene set $\mathcal{G}$, we are given a free-text annotation $y = (y_1, \dots, y_T)$ describing the biological function, pathway membership, or functional role. These annotations are concise, domain-specific phrases or sentences designed to capture the essential functional properties of the gene set without extensive narrative context.

\textbf{Task}: Given a gene set $\mathcal{G} = \{g_1, \dots, g_M\}$, with each gene $g_i$ associated with proteins
$p_{i,1}, \dots, p_{i,K_i}$ represented as amino acid sequences,
the goal of the gene set annotation task is to predict the corresponding annotation sequence
\[
y = (y_1, y_2, \dots, y_T),
\]
where $y_t \in \mathcal{V}$ are tokens drawn from the vocabulary $\mathcal{V}$ of the pretrained language model. Formally, the task can be framed as conditional sequence generation; we seek a model $f_\theta$ such that
\[
f_\theta(\mathcal{G}) = \hat{y} \approx y,
\]
where $\hat{y} = (\hat{y}_1, \hat{y}_2, \dots, \hat{y}_T)$ is the predicted annotation sequence. In other words, the model must learn a mapping from the hierarchical representation of the gene set, incorporating amino acid sequences for all constituent proteins, to a concise, technical description of the gene set’s biological function.
\subsection{Model Overview}
The model consists of two main components:

\noindent $\bullet$ A \textbf{Hierarchical Attention-Based Gene Set Encoder}, which learns context-aware gene set representations by encoding proteins with a frozen pretrained protein language model (PLM) and hierarchically aggregating them into gene- and gene set–level embeddings using attention pooling.

\noindent $\bullet$ The \textbf{Hybrid Gene Set Soft Prompting} module, wherein the resulting gene set embedding is projected by a trainable multilayer perceptron (MLP) into soft prompt tokens that condition a frozen language model for free-text annotation generation. These soft prompts are combined with task instructions and auxiliary contextual information from a large pretrained language model and fed to a local LLM to produce the final predicted annotation.
\begin{figure}[t]
    \centering
    \includegraphics[width=0.9\columnwidth]{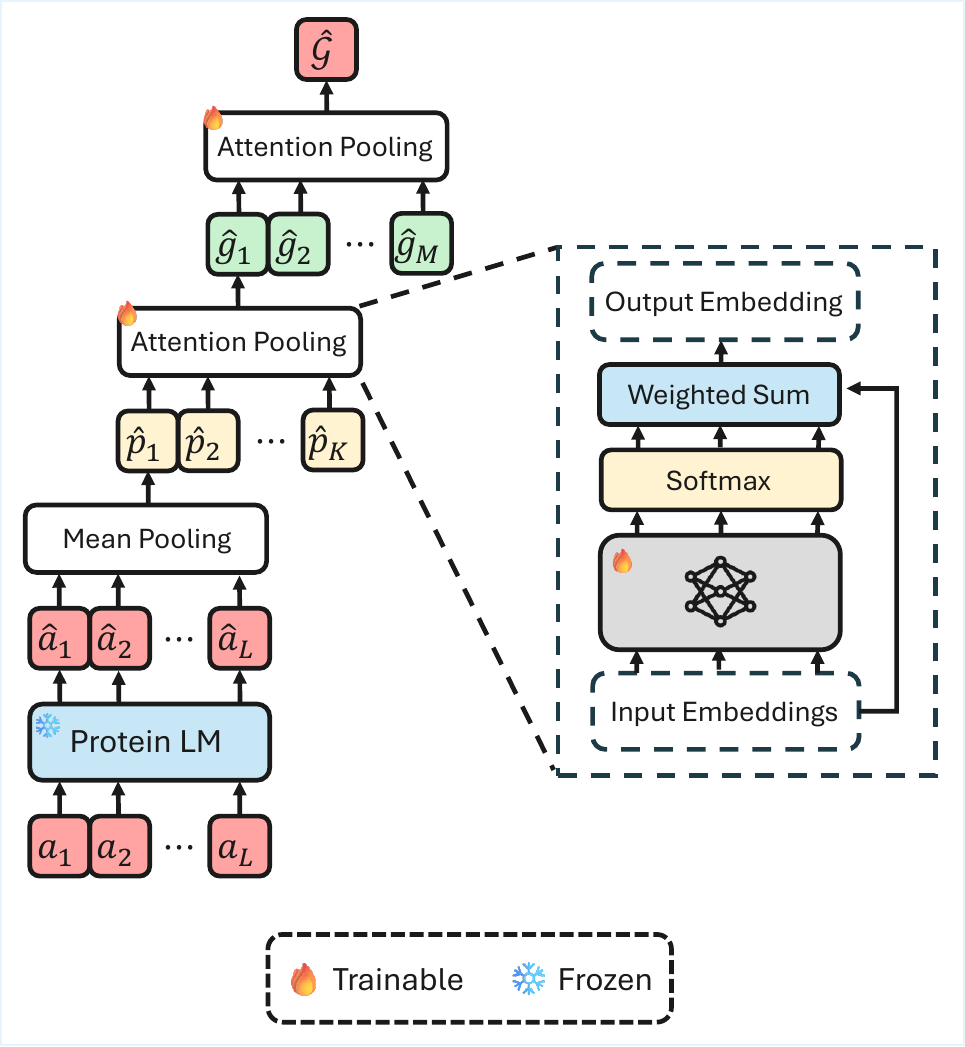}
    \caption{Overview of the hierarchical gene set encoder. Frozen pretrained protein representations are hierarchically aggregated using trainable attention-based pooling to produce a gene set embedding.}
    \label{fig:encoder}
\end{figure}
\subsection{Hierarchical Attention-Based Gene Set Encoder}
The hierarchical attention-based gene set encoder is responsible for generating gene set-level representations using trainable attention pooling mechanisms. As shown in Figure \ref{fig:encoder}, we begin by taking as input a gene set $\mathcal{G}$ comprising genes $g_1, g_2, \dots, g_M$ where $M = \lvert\mathcal{G}\rvert$. Each gene $g_i$ encodes proteins $p_1^{(i)}, p_2^{(i)}, \dots, p_{K_i}^{(i)}$, where $K_i=\lvert g_i \rvert$, and each protein $p_j^{(i)}$ is expressed as a sequence of amino acids $a_1^{(i, j)}, a_2^{(i, j)}, \dots, a_{L_{i,j}}^{i, j}$, where $L_{i, j}=\lvert p_j^{(i)}\rvert$.
Each protein $p_j^{(i)}$ is first encoded using a pretrained protein language model (ESM-2), which produces a contextual embedding for each amino acid in the sequence:
\begin{equation}
\{\hat{a}_k^{(i,j)}\}_{k=1}^{L_{i,j}} = \text{ESM}(\{a_k^{(i,j)}\}_{k=1}^{L_{i,j}})
\end{equation}
where $\hat{a}_k^{(i,j)}\in \mathbb{R}^{d}$, and $d$ is the hidden dimension of the protein language model.
To obtain a fixed-size embedding for each protein, we apply mean pooling over its amino acid embeddings. Specifically, for protein $p_j^{(i)}$, we compute:
\begin{equation}
\hat{p}_j^{(i)} = \frac{1}{L_{i,j}} \sum_{k=1}^{L_{i,j}} \hat{a}_k^{(i,j)}.
\end{equation}
For higher-level aggregation, we use attention pooling. Given a set of input embeddings $\{x_1, \ldots, x_L\}$, attention pooling is defined as follows. Each embedding is first scored using a trainable multilayer perceptron (see 
Appendix~\ref{sec:hyperparameters} for details):
\begin{equation}
h_k = \text{MLP}(x_k),
\end{equation}
The weights $\alpha_k$ are obtained via a softmax:
\begin{equation}
\alpha_k = \frac{\exp(h_k)}{\sum_{\ell=1}^{L} \exp(h_\ell)},
\end{equation}
and the pooled representation $\hat{x}$ is computed as a weighted sum:
\begin{equation}
\hat{x} = \sum_{k=1}^{L} \alpha_k x_k.
\end{equation}
We denote this operator as $\text{AttentionPooling}(\cdot)$, which is reused for multiple aggregation steps across different levels of the hierarchy.
Given the set of protein embeddings associated with a gene $g_i$, corresponding to its annotated splice isoforms, we aggregate them into a single gene-level representation using the $\text{AttentionPooling}(\cdot)$ operator defined above. Specifically, we apply attention pooling to $\{\hat{p}_j^{(i)}\}_{j=1}^{K_i}$ with a gene-level pooling MLP to obtain the gene embedding $\hat{g}_i$:
\begin{equation}
\hat{g}_i = \text{AttentionPooling}_{\text{gene}}\left(\{\hat{p}_j^{(i)}\}_{j=1}^{K_i}\right).
\end{equation}
This allows the model to differentially weight splice isoforms according to their relevance, 
downweighting noisy or irrelevant isoforms and producing a gene representation that reflects 
isoform-specific contributions to be used as input to the subsequent gene set--level 
aggregation.


\begin{figure}[t]
    \centering
    \includegraphics[width=\columnwidth]{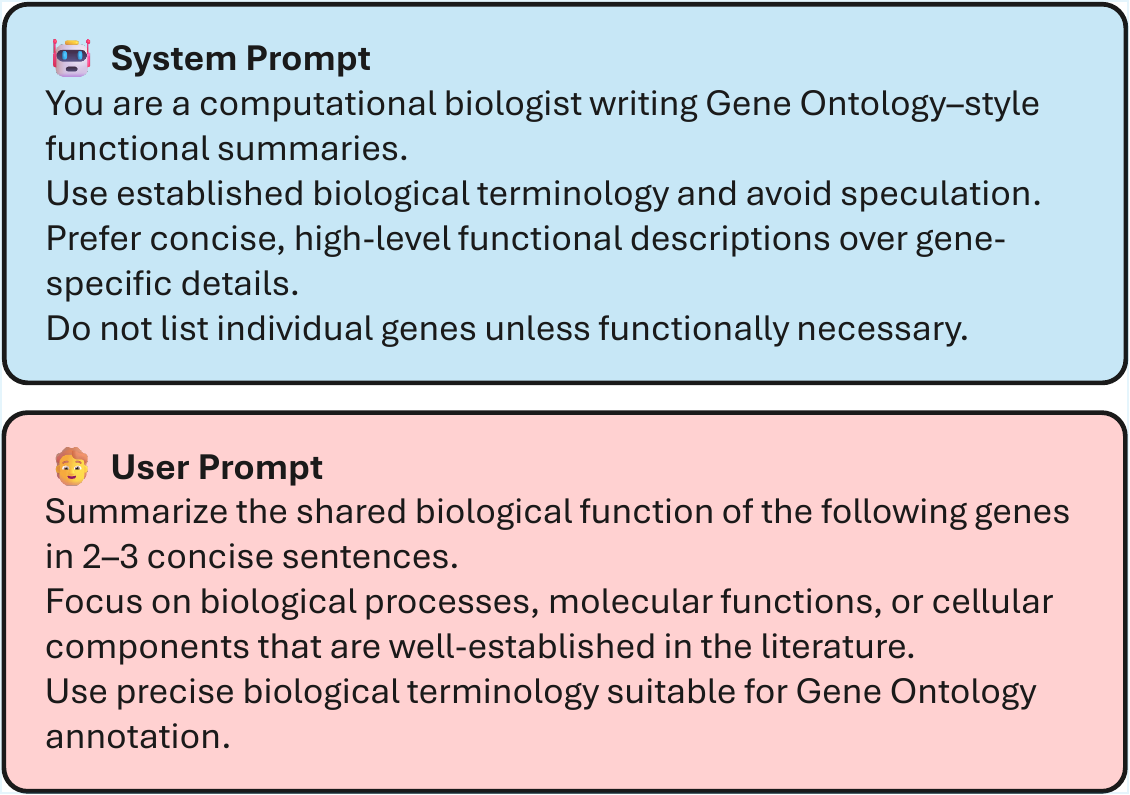}
    \caption{To obtain the gene set context, we utilize a system prompt to define a persona and stylistic guidelines
alongside a user prompt for task instructions, which are
fed to a commercial model.}
    \label{fig:prompts}
\end{figure}
Given a gene set $\mathcal{G}$ with gene embeddings $\{\hat{g}_i\}_{i=1}^{M}$, we apply the $\text{AttentionPooling}(\cdot)$ operator with independently learned parameters to produce a fixed-size gene set representation $\hat{\mathcal{G}}$:
\begin{equation}
\hat{\mathcal{G}} = \text{AttentionPooling}_{\text{set}}\left(\{\hat{g}_i\}_{i=1}^{M}\right).
\end{equation}
This final representation captures the relative contributions of genes within the set and serves as input to downstream components.

\begin{figure}[t]
    \centering
    \includegraphics[width=0.85\columnwidth]{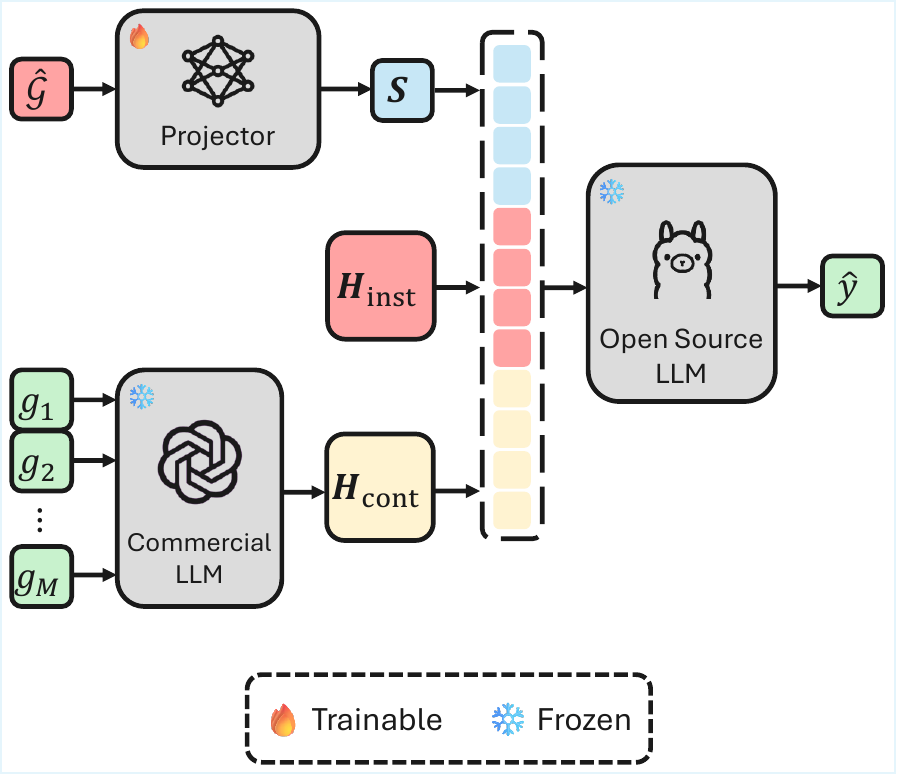}
    \caption{Overview of the hybrid gene set soft prompting module. A gene set embedding is converted into soft prompt tokens that, together with instruction and context prompts, condition a frozen language model to generate free-text gene set annotations.}
    \label{fig:soft_prompting_module}
\end{figure}

\subsection{Hybrid Gene Set Soft Prompting}
As shown in Figure \ref{fig:soft_prompting_module}, this module generates free-text annotations from the gene set embedding produced by the hierarchical encoder.

To condition the local LLM (e.g., Llama) on the gene set embedding, we map it to the LLM input space using a multi-layer perceptron (MLP):
\begin{equation}
\boldsymbol{S} = \text{MLP}_{\text{soft}}(\hat{\mathcal{G}}),
\end{equation}
where $\boldsymbol{S} \in \mathbb{R}^{n \times d_\text{LLM}}$ is a sequence of $n$ tunable soft tokens in the embedding space of the LLM. These tokens serve as a learnable continuous prompt that encodes the structured biological information from the gene set.

In addition to the soft prompt, we append two hard prompts to guide the LLM during generation. The first prompt, $\boldsymbol{H}_{inst}$, contains the instruction for the model, specifying that it should generate a candidate annotation for the input gene set. The second prompt, $\boldsymbol{H}_{cont}$, is generated by a commercial large pretrained language model\footnote{This component can be replaced with any sufficiently capable locally-hosted language model for fully offline deployment.} (in our case, GPT 4.1 Nano) and provides context and analysis of the genes in the set through semantic-level reasoning as follows:
\begin{equation}
    \boldsymbol{H}_{cont} = \text{LLM}_{\text{commercial}}(\left[g_1, g_2, \dots, g_M\right]).
\end{equation}
The prompts passed to $\text{LLM}_{commercial}$ are shown in Figure \ref{fig:prompts}. This context supplies knowledge that the smaller local LLM may not encode in its pretrained weights.

The final input to the LLM (a local Llama 3 model \cite{grattafiori2024llama}) is the concatenation of the instruction, context, and soft prompts, which is then used to generate the candidate gene set annotation:
\begin{equation}
\hat{y}=\text{LLM}_\text{local}\left(\left[\boldsymbol{S}; \boldsymbol{H}_{inst}; \boldsymbol{H}_{cont}\right]\right).
\end{equation}

\vspace{-0.3cm}
\subsection{Training Procedure}

Given the ground-truth annotation sequence $y = (y_1, \dots, y_T)$, the language model defines a conditional distribution over output tokens:
\begin{equation}
P(y \mid \hat{\mathcal{G}}) = \prod_{t=1}^{T} P\left(y_t \mid y_{<t}, \boldsymbol{S}, \boldsymbol{H}_{\text{inst}}, \boldsymbol{H}_{\text{cont}}\right).
\end{equation}

Training is performed using teacher forcing, where at each time step the model conditions on the ground-truth prefix $y_{<t}$. The training objective is the standard token-level cross-entropy loss:
\begin{equation}
\mathcal{L}_{\text{LM}} = - \sum_{t=1}^{T} \log P\left(y_t \mid y_{<t}, \boldsymbol{S}, \boldsymbol{H}_{\text{inst}}, \boldsymbol{H}_{\text{cont}}\right).
\end{equation}

During optimization, the parameters of the PLM and language model are kept frozen. Gradients are propagated only through the soft prompt projector $\text{MLP}_{\text{soft}}$ and the hierarchical attention encoder, allowing the model to learn an effective conditioning of the language model on gene set representations while fully leveraging pretrained knowledge.

%% file: sections/experiments.tex
\section{Experiments}
\vspace{-0.05cm}
\subsection{Datasets}
 We evaluate our framework on two widely used gene set databases. Statistics for each dataset after filtering are presented in Table \ref{tab:statistics}. The datasets are elaborated as follows:

\noindent $\bullet$ \textbf{Gene Ontology (GO):} The Gene Ontology (GO) dataset \cite{ashburner2000gene} provides a structured, controlled vocabulary for describing gene functions across three domains: Biological Process (BP), Molecular Function (MF), and Cellular Component (CC).

\noindent $\bullet$ \textbf{MSigDB:}
The Molecular Signatures Database (MSigDB) \cite{liberzon2015molecular} is a comprehensive collection of annotated gene sets. We use gene sets from the C2 collection, specifically from KEGG \cite{kanehisa2000kegg}, Reactome \cite{fabregat2018reactome}, PID \cite{schaefer2009pid}, and BioCarta \cite{nishimura2001biocarta}.

\begin{table}[t]
  \caption{Dataset statistics (annotation length measured in characters).}
  \label{tab:statistics}
  \centering
  \small
  \begin{tabular}{lcc}
    \toprule
    \textbf{Measure}&\textbf{GO}&\textbf{MSigDB}\\
    \midrule
    Number of Gene Sets & 18,889 & 2,461 \\
    Avg. Genes Per Set & 21.18 & 50.07 \\
    Min. Number of Genes & 1 & 5 \\
    Max. Number of Genes & 13,616 & 1,491 \\
    Unique Genes Covered & 42,597 & 12,021 \\
    Avg. Annotation Length & 38.51 & 36.25 \\
    Min. Annotation Length & 4 & 6 \\
    Max. Annotation Length & 187 & 155 \\
  \bottomrule
\end{tabular}
\end{table}
\subsection{Experimental Settings}
We publicly release our source code for reproducibility.\footnote{Code: \url{https://github.com/KU-AI4H/softgene}.} We randomly split the data into 70\% training and 30\% testing sets and report the mean and 95\% confidence interval over three runs. Hyperparameter settings for both modules are presented in Appendix C.1.
\begin{table*}[ht]
\centering
\caption{Performance of SoftGene and baselines on the Gene Ontology (GO) and MSigDB datasets. First, second, and third-best results are indicated by \textbf{boldface}, \uuline{double underline}, and \uline{single underline}, respectively.}
\label{tab:baselines}
\small
\begin{tabular}{lcccccc}
\toprule
\multirow{2}{*}{\textbf{Configuration}} & \multicolumn{3}{c}{\textbf{Gene Ontology (GO)}} & \multicolumn{3}{c}{\textbf{MSigDB}} \\
\cmidrule(lr){2-4} \cmidrule(lr){5-7}
 & METEOR & ROUGE-L & BERTScore-F1 & METEOR & ROUGE-L & BERTScore-F1 \\
\midrule
\multicolumn{7}{c}{\textbf{Prompt Engineering Methods}} \\
\midrule
Zero-shot    & 0.024 \tiny{± 0.001}  & 0.133 \tiny{± 0.001} & 0.773 \tiny{± 0.00} & 0.087 \tiny{± 0.014}  & 0.127 \tiny{± 0.009} & 0.762 \tiny{± 0.002} \\
Few-shot & 0.018 \tiny{± 0.001} & 0.164 \tiny{± 0.003} & 0.740 \tiny{± 0.001} & 0.100 \tiny{± 0.025} & 0.136 \tiny{± 0.017} & 0.771 \tiny{± 0.008} \\
One-shot CoT & 0.065 \tiny{± 0.001} & 0.096 \tiny{± 0.002} & 0.761 \tiny{± 0.001} & 0.086 \tiny{± 0.010} & 0.112 \tiny{± 0.003} & 0.762 \tiny{± 0.001} \\
\midrule
\multicolumn{7}{c}{\textbf{Prompt-tuning Methods}} \\
\midrule
Gene Symbol Soft Prompting & 0.113 \tiny{± 0.002} & 0.158 \tiny{± 0.004} & 0.778 \tiny{± 0.001} & 0.105 \tiny{± 0.022} & 0.138 \tiny{± 0.021} & 0.771 \tiny{± 0.008} \\
P-Tuning    & 0.148 \tiny{± 0.130} & 0.196 \tiny{± 0.130} & 0.783 \tiny{± 0.042} & 0.113 \tiny{± 0.058} & 0.161 \tiny{± 0.061} & 0.716 \tiny{± 0.026} \\
Prefix Tuning    & 0.142 \tiny{± 0.082} & 0.173 \tiny{± 0.094} & 0.766 \tiny{± 0.048} & 0.080 \tiny{± 0.006}  & 0.113 \tiny{± 0.021} & 0.756 \tiny{± 0.007} \\
\midrule
\multicolumn{7}{c}{\textbf{Agentic Systems}} \\
\midrule
Discovera & 0.107 \tiny{± 0.003} & 0.144 \tiny{± 0.005} & 0.769 \tiny{± 0.001} & 0.092 \tiny{± 0.003} & 0.108 \tiny{± 0.003} & 0.768 \tiny{± 0.007} \\
TALISMAN & 0.108 \tiny{± 0.006} & 0.146 \tiny{± 0.002} & 0.766 \tiny{± 0.001} & 0.086 \tiny{± 0.005} & 0.107 \tiny{± 0.001} & 0.760 \tiny{± 0.001} \\
GeneAgent & 0.071 \tiny{± 0.003} & 0.116 \tiny{± 0.002} & 0.768 \tiny{± 0.001} & 0.030 \tiny{± 0.006} & 0.117 \tiny{± 0.003} & 0.764 \tiny{± 0.002} \\
\midrule
\multicolumn{7}{c}{\textbf{SoftGene}} \\
\midrule
SoftGene (Mistral 3 8B) & \uline{0.203 \tiny{± 0.015}} & \uline{0.257 \tiny{± 0.015}} & \uline{0.802 \tiny{± 0.001}} & \uuline{0.148 \tiny{± 0.017}} & \uline{0.206 \tiny{± 0.018}} & \uuline{0.795 \tiny{± 0.005}}\\
SoftGene (Qwen 3 8B) & \uuline{0.210 \tiny{± 0.007}} & \uuline{0.260 \tiny{± 0.003}} & \uuline{0.803 \tiny{± 0.001}} & 0.135 \tiny{± 0.023} & 0.194 \tiny{± 0.021} & 0.790 \tiny{± 0.007} \\
SoftGene (Llama 3.2 3B) & 0.197 \tiny{± 0.013} & 0.247 \tiny{± 0.003} & 0.800 \tiny{± 0.001} & \uline{0.145 \tiny{± 0.004}} & \uuline{0.206 \tiny{± 0.011}} & \uline{0.793 \tiny{± 0.004}} \\
SoftGene (Llama 2 13B) & 0.191 \tiny{± 0.008} & 0.239 \tiny{± 0.011} & 0.797 \tiny{± 0.004} & 0.121 \tiny{± 0.024} & 0.184 \tiny{± 0.024} & 0.787 \tiny{± 0.009} \\
SoftGene (Llama 3.1 8B) & \textbf{0.220 \tiny{± 0.033}} & \textbf{0.267 \tiny{± 0.029}} & \textbf{0.806 \tiny{± 0.006}} & \textbf{0.157 \tiny{± 0.005}} & \textbf{0.215 \tiny{± 0.001}} & \textbf{0.795 \tiny{± 0.003}} \\
\bottomrule
\end{tabular}
\vspace{-0.2cm}
\end{table*}

\subsection{Evaluation Metrics}
Following previous works \cite{wang2024geneagentselfverificationlanguageagent, wang2025gene}, we employ lexical and semantic similarity measures to evaluate our method. We use the following metrics for our evaluation:

\noindent $\bullet$ \textbf{METEOR} \cite{10.5555/1626355.1626389} measures the alignment between the candidate annotation and the reference by considering unigram matches while accounting for synonyms, stemming, and word order. We report METEOR to capture lexical alignment while accounting for synonyms, stemming, and word order.

\noindent $\bullet$ \textbf{ROUGE} \cite{chin2004rouge} identifies the longest common subsequence between the generated and reference texts. This ensures that essential biological keywords and descriptions from the ground truth are captured in the model's output.

\noindent $\bullet$ \textbf{BERTScore} \cite{zhang2019bertscore} addresses the limitations of exact-match metrics by using contextual embeddings to calculate similarity. By comparing tokens based on their vector representations, BERTScore rewards candidate annotations that are synonymous with the ground truth, even when exact phrasing differs.
\subsection{Results}
\vspace{-0.05cm}
\subsubsection{Baseline Comparison }
We compare our approach against the following categories of baselines, as summarized in Table \ref{tab:baselines}.

\noindent $\bullet$ \textbf{Prompt Engineering Methods}: We evaluate standard zero-shot and few-shot prompting \mbox{\cite{brown2020language}}, as well as a one-shot chain-of-thought (CoT) prompting strategy following prior work \cite{hu2025evaluation}. These baselines rely entirely on the language model's knowledge acquired during pretraining. Our method consistently outperforms them across all metrics, indicating the limitations of prompting alone for this task.

\noindent $\bullet$ \textbf{Prompt-tuning Methods}: We compare against two standard prompt-tuning techniques, P-Tuning \cite{liu2022ptuningv2prompttuning} and Prefix Tuning \cite{li2021prefixtuningoptimizingcontinuousprompts}, as well as a baseline that uses token embeddings averaged across all gene symbols in the set as a soft prompt. These methods allow the model to adapt to the task with limited additional parameters. These prompt-tuning baselines underperform SoftGene under our evaluation setting, resulting in lower performance than our approach.

\noindent $\bullet$ \textbf{Agentic Systems}: We evaluate agentic baselines that incorporate external knowledge sources. Discovera \cite{veizaga2025gene} and TALISMAN \cite{joachimiak2024genesetsummarizationusing} augment generation with database queries, while GeneAgent \cite{wang2024geneagentselfverificationlanguageagent} combines retrieval-augmented generation (RAG) with self-reflection mechanisms to produce grounded outputs. Despite access to external biological databases, these methods underperform our model, showing that these agentic baselines do not match SoftGene under our evaluation setting.
\begin{table*}
  \centering
  \caption{Ablation study.}
  \vspace{-0.2cm}
  \label{tab:ablations}
  \small
  \begin{tabular}{lcccccc}
    \toprule
    \multirow{2}{*}{\textbf{Configuration}} & \multicolumn{3}{c}{\textbf{Gene Ontology (GO)}} & \multicolumn{3}{c}{\textbf{MSigDB}} \\
    \cmidrule(lr){2-4} \cmidrule(lr){5-7}
    & METEOR & ROUGE-L & BERTScore & METEOR & ROUGE-L & BERTScore \\
    \midrule
    w/o Soft Prompt ($S$)    & 0.020 \tiny{± 0.000}  & 0.065 \tiny{± 0.001} & 0.707 \tiny{± 0.001} & 0.006 \tiny{± 0.004} & 0.045 \tiny{± 0.002} & 0.704 \tiny{± 0.001}\\
    w/o Context ($H_{cont}$) & 0.182 \tiny{± 0.024} & 0.232 \tiny{± 0.028} & 0.795 \tiny{± 0.005} & 0.145 \tiny{± 0.007} & 0.202 \tiny{± 0.020} & 0.790 \tiny{± 0.003} \\
    w/o Protein Embeddings    & 0.199 \tiny{± 0.015} & 0.252 \tiny{± 0.017} & 0.803 \tiny{± 0.002} & 0.097 \tiny{± 0.017} & 0.135 \tiny{± 0.025} & 0.774 \tiny{± 0.012} \\
    \textbf{SoftGene}        & 0.220 \tiny{± 0.033} & 0.267 \tiny{± 0.029} & 0.806 \tiny{± 0.006} & 0.157 \tiny{± 0.005} & 0.215 \tiny{± 0.001} & 0.795 \tiny{± 0.003} \\
    \bottomrule
  \end{tabular}
\end{table*}
\begin{figure*}[t!]
    \centering
    \includegraphics[width=\textwidth]{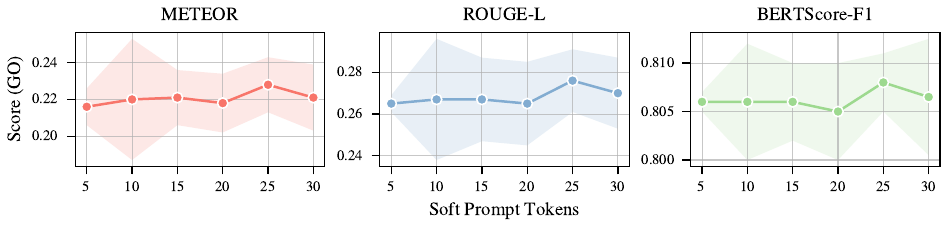}
    \vspace{-0.5cm}
    \caption{Performance comparison of the SoftGene model across different soft prompt lengths using the GO dataset.}
    \label{fig:go_soft_prompt}
\end{figure*}
\begin{figure*}[t!]
    \centering
    \includegraphics[width=\textwidth]{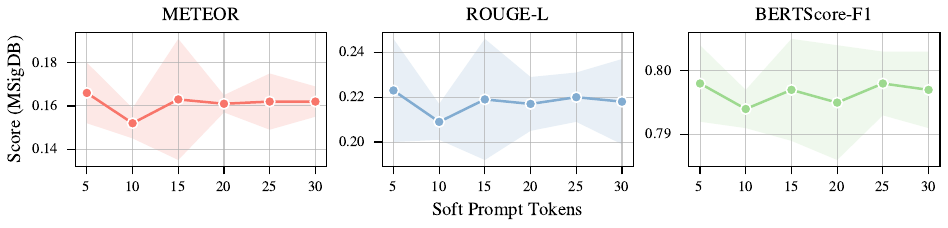}
    \vspace{-0.5cm}
    \caption{Performance comparison of the SoftGene model across different soft prompt lengths using the MSigDB dataset.}
    \vspace{-0.3cm}
    \label{fig:msigdb_soft_prompt}
\end{figure*}

\subsubsection{Ablation Study}
\vspace{-0.2cm}
We conduct an ablation study in Table \ref{tab:ablations} to evaluate the contribution of each component in SoftGene.
Removing the soft prompt and relying solely on hard prompts (“w/o Soft Prompt ($S$)'') leads to a substantial performance drop across all metrics and both datasets, highlighting the importance of directly conditioning the LLM on structured gene set representations.
Excluding LLM-generated contextual information (“w/o Context ($\text{H}_{cont}$)”) results in a consistent degradation in performance while still outperforming baselines, indicating that high-level semantic context improves generation quality.
When replacing pretrained PLM embeddings with randomly initialized embeddings (‘w/o Protein Embeddings’), overall performance decreases, indicating that sequence-derived representations provide complementary information beyond the downstream architecture. As shown by the per-domain analysis, however, the magnitude of this contribution varies substantially across biological domains.
Notably, without pretrained protein embeddings, SoftGene still outperforms all baselines, demonstrating that gains are not simply attributable to PLM embeddings.
These results confirm that SoftGene’s gains arise from the combined effect of structured gene set embeddings, attention-based pooling, and soft prompt conditioning, with each contributing meaningfully to final performance.
\subsubsection{Soft Prompt Length}
We analyze the effect of soft prompt length on model performance by varying the number of learned 
soft prompt tokens and evaluating outputs. Results for the GO and MSigDB 
datasets are shown in Figures \ref{fig:go_soft_prompt} and \ref{fig:msigdb_soft_prompt}. Across both datasets, performance varies only modestly across the evaluated prompt lengths.
Since increasing prompt length 
yields no significant performance gains, we select 10 tokens as an empirical choice to reduce computational cost while preserving information.

\begin{table*}[t]
\centering
\caption{Performance comparison between random and gene-disjoint train-test splits.}
\label{tab:gene_disjoint_results}
\small
\begin{tabular}{lcccccc}
\toprule
& \multicolumn{3}{c}{\textbf{Gene Ontology (GO)}} & \multicolumn{3}{c}{\textbf{MSigDB}} \\
\cmidrule(lr){2-4} \cmidrule(lr){5-7}
\textbf{Split} & METEOR & ROUGE-L & BERTScore & METEOR & ROUGE-L & BERTScore \\
\midrule
Gene-Disjoint
& 0.205 \tiny{± 0.014}
& 0.251 \tiny{± 0.015}
& 0.801 \tiny{± 0.009}
& 0.136 \tiny{± 0.082}
& 0.193 \tiny{± 0.062}
& 0.791 \tiny{± 0.034}
\\

Random
& 0.220 \tiny{± 0.033}
& 0.267 \tiny{± 0.029}
& 0.806 \tiny{± 0.006}
& 0.157 \tiny{± 0.005}
& 0.215 \tiny{± 0.001}
& 0.795 \tiny{± 0.003}
\\
\bottomrule
\end{tabular}
\end{table*}
\subsubsection{Gene-Disjoint Generalization}

To further evaluate whether SoftGene learns biologically meaningful sequence representations rather than memorizing gene co-occurrence patterns, we conduct additional experiments using a gene-disjoint train-test split. In this setting, train and test partitions are constructed such that genes appearing in the test set are excluded from the training set as much as possible while maintaining an approximate 70/30 split ratio. This provides a substantially more challenging evaluation setting by requiring the model to generalize to previously unseen genes rather than relying on overlap between training and evaluation gene sets. Results are reported in Table~\ref{tab:gene_disjoint_results}. We observe that SoftGene trained under the gene-disjoint split performs comparably to the standard random split across all evaluation metrics on both the GO and MSigDB datasets. While modest decreases in performance are expected due to the stricter evaluation protocol, the overall degradation remains limited. These findings suggest that SoftGene is not solely memorizing recurring gene compositions, but instead learns transferable biological sequence patterns and functional relationships that generalize to unseen genes.

\begin{table*}[t]
\centering
\caption{Per-domain ablation of PLM embeddings on the GO dataset. Comparison between real PLM embeddings and randomized embeddings, with differences.}
\resizebox{\textwidth}{!}{%
\label{tab:domain_ablation}
\small
\setlength{\tabcolsep}{5pt}
\renewcommand{\arraystretch}{1.1}
\begin{tabular}{llccc ccc ccc}
\toprule
& \multicolumn{3}{c}{\textbf{PLM}} 
& \multicolumn{3}{c}{\textbf{Randomized}} 
& \multicolumn{3}{c}{\textbf{$\Delta$}} \\
\cmidrule(lr){2-4} \cmidrule(lr){5-7} \cmidrule(lr){8-10}
\textbf{Domain} 
& METEOR & ROUGE-L & BERTScore
& METEOR & ROUGE-L & BERTScore
& METEOR & ROUGE-L & BERTScore \\
\midrule
BP
& 0.223 & 0.264 & 0.802
& 0.222 & 0.256 & 0.798
& -0.001 & -0.008 & -0.004 \\

CC
& 0.164 & 0.212 & 0.784
& 0.102 & 0.151 & 0.769
& -0.062 & -0.061 & -0.015 \\

MF
& 0.236 & 0.314 & 0.825
& 0.213 & 0.285 & 0.820
& -0.023 & -0.029 & -0.005 \\
\bottomrule
\end{tabular}}
\vspace{-0.3cm}
\end{table*}
\subsubsection{Per-Domain PLM Embedding Ablation}
To better understand the contribution of protein language model embeddings within the SoftGene framework, we present a fine-grained ablation broken down by GO domain (MF, BP, CC) in Table \ref{tab:domain_ablation}. For each domain, we compare SoftGene with real PLM embeddings against a randomized embedding baseline, where protein embeddings are replaced with random noise of the same dimensionality. This isolates the domains in which protein sequence information provides the greatest benefit.
Notably, the effect of PLM embeddings varies across domains, with larger improvements observed in CC compared to BP and MF, suggesting that different GO domains may differentially benefit from sequence-derived representations.
\vspace{-6pt}
\subsubsection{Case Verification: Zinc Ion Transmembrane Transport}
We first examine the gene set GO:0071577 (zinc ion transmembrane transport), for which the model produces a near-perfect prediction. The highest attention weights are assigned to genes from known zinc transporter families, including SLC39 (ZIP) and SLC30 (ZnT), which mediate zinc influx and efflux across cellular membranes \cite{jeong2013slc39, eide2006zinc}. The top-ranked genes, SLC39A5 and SLC39A4, are well-characterized zinc importers \cite{wang2004mammalian}; SLC39A4 is mutated in acrodermatitis enteropathica \cite{wang2002novel}, a classic inherited zinc deficiency disorder. Genes less directly related to zinc transport, such as the glutamate transporter SLC1A1 \cite{kanai2003glutamate}, receive lower attention weights. This indicates that the gene-level attention layer ranks genes by functional relevance rather than treating all set members equally.
\subsubsection{Interpretability: Smooth Muscle Signaling}
\begin{figure}[t!]
    \centering
    \includegraphics[width=0.5\textwidth]{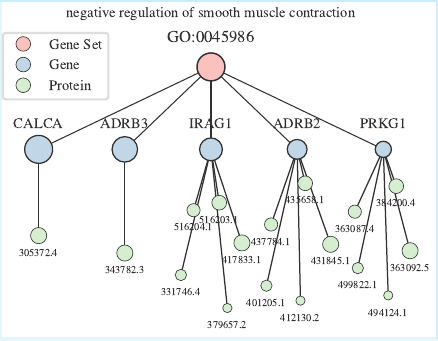}
    \caption{Attention visualization for GO:0045986. The tree depicts the three-level hierarchy of the gene set encoder, with the top five protein isoforms (by attention weight) shown beneath each gene. Node size reflects learned attention weights, indicating the relative contribution of each gene and protein isoform.}
    \label{fig:attn}
\end{figure}

We next analyze attention scores for GO:0045986 (negative regulation of smooth muscle contraction), shown in Figure \ref{fig:attn}. SoftGene predicts the related term “negative regulation of cGMP biosynthetic process,” consistent with the nitric oxide–cGMP pathway that regulates vascular smooth muscle relaxation \cite{hofmann2006function}.
At the gene level, the highest attention weights are assigned to CALCA, encoding the vasodilator calcitonin gene-related peptide \cite{brain2004vascular}, and the beta-adrenergic receptors ADRB2 and ADRB3, which promote smooth muscle relaxation via cAMP signaling \cite{johnson2006molecular}.
At the isoform level, the highest-weighted PRKG1 protein is ENSP00000363092.5 (PRKG1$\alpha$). Unlike PRKG1$\beta$, PRKG1$\alpha$ localizes to smooth muscle stress fibers and mediates vascular relaxation in response to cGMP \cite{francis2010cgmp, hofmann2006function}. This preference emerges without isoform-level supervision, suggesting that protein language model embeddings encode sequence signals distinguishing functional splice variants.

%% file: sections/relatedworks.tex
\section{Related Works}
\textbf{Automatic Gene Set Annotation:}
Recent evaluations show that large language models outperform traditional methods for gene set annotation by synthesizing knowledge from biomedical text, while remaining susceptible to hallucinated biological associations \cite{hu2025evaluation}.  
TALISMAN evaluates instruction-tuned LLMs for gene set function summarization and finds that their outputs remain dependent on curated biological knowledge bases to ensure reliability \cite{joachimiak2024genesetsummarizationusing}.  
GeneAgent introduces a self-verifying language agent for gene set knowledge discovery, using iterative verification to improve factual consistency \cite{wang2024geneagentselfverificationlanguageagent}.  
Discovera integrates LLMs with retrieval-based biological tools to ground gene set annotations in external knowledge sources and mitigate hallucinations \cite{veizaga2025gene}.  

\noindent\textbf{LLM Soft Prompting:}
Advances in parameter-efficient fine-tuning have led to a family of methods that steer frozen models through learned prompts rather than updating model weights.
Prefix-Tuning introduces continuous, learnable vectors prepended to Transformer layers, enabling parameter-efficient steering of frozen models \cite{li2021prefixtuningoptimizingcontinuousprompts}.  
Prompt Tuning shows that, for sufficiently large models, optimizing soft tokens at the input embedding layer can achieve performance comparable to full fine-tuning \cite{lester2021powerscaleparameterefficientprompt}.  
P-Tuning models dependencies among soft prompts using a continuous prompt encoder, improving stability and allowing prompts to be conditioned on specific inputs \cite{liu2024gpt}.  
P-Tuning v2 extends soft prompting to deeper Transformer layers to maintain effectiveness on smaller-scale models \cite{liu2022ptuningv2prompttuning}.  
Soft prompting has also been explored in biological domains, including multimodal protein modeling in ProCyon \cite{queen2024procyon}, gene-focused dialogue systems in GeneChat \cite{dhanasekar2025genechat}, and genomic reasoning such as BioReason \cite{fallahpour2025bioreason}.  

%% file: sections/conclusion.tex
\section{Conclusion}
In this work, we presented SoftGene, a hybrid prompting framework that integrates a hierarchical attention-based encoder over protein sequences with soft prompt conditioning for language model--driven gene set annotation. Experimental results on the Gene Ontology and MSigDB benchmarks show that SoftGene outperforms prompt-based, agent-based, and prompt-tuning baselines across the reported metrics. Additional analyses show that protein-sequence representations provide complementary information, while their contribution varies substantially across biological domains. These findings support sequence-aware soft prompting as a promising approach for interpretable gene set annotation while highlighting the importance of domain-specific evaluation and biological validation.

%% file: sections/appendices.tex
\appendix



\setcounter{topnumber}{4}
\setcounter{bottomnumber}{4}
\setcounter{totalnumber}{8}
\renewcommand{\topfraction}{0.95}
\renewcommand{\bottomfraction}{0.95}
\renewcommand{\textfraction}{0.03}
\renewcommand{\floatpagefraction}{0.85}

\newenvironment{apptable}{
  \small
  \setlength{\tabcolsep}{4pt}
  \renewcommand{\arraystretch}{1.02}
}{}

\newcommand{\apptocsec}[2]{\par\medskip\noindent\textbf{\ref{#1}\enspace #2}\hfill\textbf{\pageref{#1}}\par\smallskip}
\newcommand{\apptocsub}[2]{\noindent\hspace*{1.5em}\ref{#1}\enspace #2\ \dotfill\ \pageref{#1}\par}

\vspace{0.3cm}
\section*{Appendix Contents}
\begingroup\small
\apptocsec{app:additional_experiments}{Additional Experiments}
\apptocsub{app:llm_backbone}{LLM Backbone for Context Generation}
\apptocsub{app:protein_lm}{Encoder Protein Language Model}
\apptocsub{app:description_length}{Performance by Description Length}
\apptocsub{app:cardinality}{Performance by Gene Set Cardinality}
\apptocsub{app:domain_collection}{Performance by Domain and Collection}
\apptocsub{app:gene_uniqueness}{Performance by Gene Uniqueness}
\apptocsub{app:ontology_depth}{Performance by Ontology Depth}
\apptocsub{sec:additional_metrics}{Additional Performance Metrics}
\apptocsub{app:case_studies}{Additional Case Studies}

\apptocsec{app:qualitative_analysis}{Qualitative Analysis}
\apptocsub{app:high_low_completions}{High- and Low-Scoring Completions}
\apptocsub{app:failure_analysis}{Failure Mode Analysis}

\apptocsec{app:implementation_details}{Implementation Details}
\apptocsub{sec:hyperparameters}{Hyperparameters}
\apptocsub{app:baselines}{Baselines}
\apptocsub{app:runtime_analysis}{Runtime Analysis}

\apptocsec{app:dataset_preparation}{Dataset Preparation}
\apptocsub{app:protein_sequences}{Protein Sequences and Isoforms}
\apptocsub{app:go_dataset}{GO Dataset}
\apptocsub{app:msigdb_dataset}{MSigDB Dataset}
\apptocsub{app:contamination}{Dataset Contamination Analysis}

\apptocsec{app:llm_prompts}{LLM Prompts}
\apptocsub{app:softgene_prompt}{SoftGene Prompt}
\apptocsub{app:zeroshot_prompt}{Zero-Shot Prompt}
\apptocsub{app:fewshot_prompt}{Few-Shot Prompt}
\apptocsub{app:cot_prompt}{One-Shot Chain-of-Thought Prompt}

\apptocsec{app:related_works}{Additional Related Works}
\apptocsub{app:protein_language_models}{Protein Language Models}
\apptocsub{app:traditional_gene_set}{Traditional Gene Set Annotation Methods}

\endgroup


\vspace{0.7cm}
\section{Additional Experiments}
\label{app:additional_experiments}
\subsection{LLM Backbone for Context Generation}
\label{app:llm_backbone}
\begin{figure*}[t]
    \centering
    \includegraphics[width=0.9\linewidth]{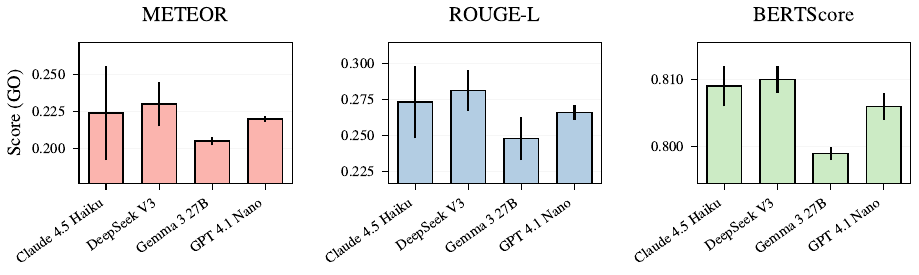}
    \caption{Performance comparison of SoftGene across different commercial models using the GO dataset.}
    \label{fig:commercial_ablation1}
\end{figure*}

\begin{figure*}[t]
    \centering
    \includegraphics[width=\linewidth]{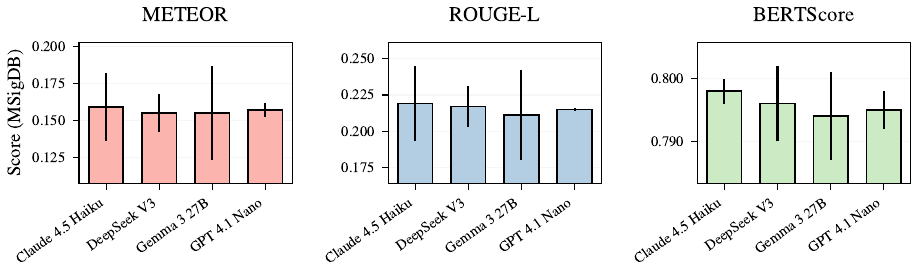}
    \caption{Performance comparison of SoftGene across different commercial models using the MSigDB dataset.}
    \label{fig:commercial_ablation2}
\end{figure*}

We evaluate the impact of different commercial large language models for generating textual context describing each gene set in Figures \ref{fig:commercial_ablation1} and \ref{fig:commercial_ablation2}. Specifically, we compare contexts produced by GPT, DeepSeek \cite{liu2024deepseek}, Claude, and Gemma \cite{team2025gemma} while keeping all other components fixed. Contexts generated by GPT and DeepSeek result in the strongest overall performance across all performance metrics, with DeepSeek achieving the highest average scores. Claude yields slightly lower but comparable performance, while Gemma consistently underperforms the other models.
\subsection{Encoder Protein Language Model}
\label{app:protein_lm}
\begin{figure*}[t]
    \centering
    \includegraphics[width=0.9\linewidth]{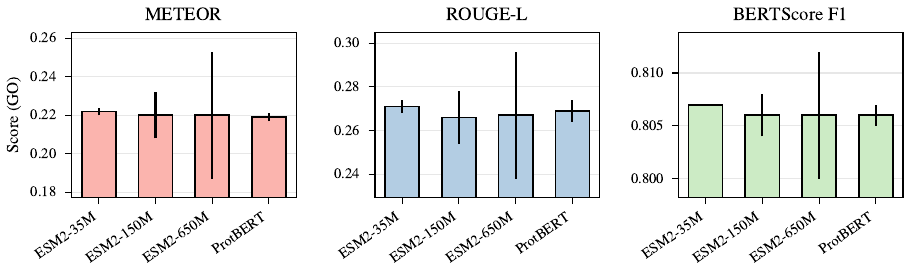}
    \caption{Performance comparison of SoftGene across different protein language models using the GO dataset.}
    \label{fig:plm_ablation1}
\end{figure*}

\begin{figure*}[t]
    \centering
    \includegraphics[width=\linewidth]{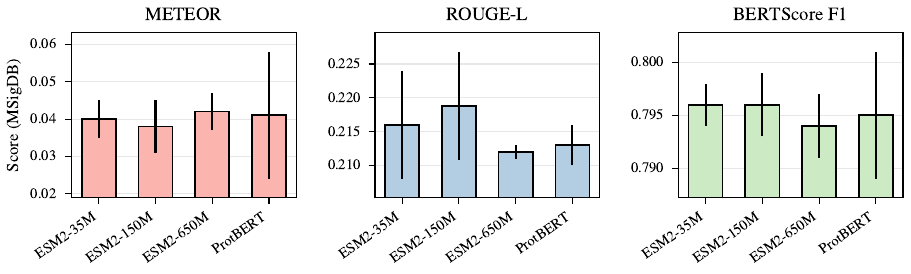}
    \caption{Performance comparison of SoftGene across different protein language models using the MSigDB dataset.}
    \label{fig:plm_ablation2}
\end{figure*}

We evaluate the robustness of our approach to the choice of protein language model by training with embeddings generated from multiple protein encoders, including three ESM-2 checkpoints of varying capacity and ProtBERT \cite{elnaggar2021prottrans}, a BERT model trained on amino acid sequence data. As shown in Figures \ref{fig:plm_ablation1} and \ref{fig:plm_ablation2}, performance is highly consistent across models, with minimal variation in performance metrics across random seeds. These results indicate that our method is not sensitive to the specific protein embedding backbone used, suggesting that improvements stem from how gene set representations are integrated into the language model rather than from any particular protein encoder.
\subsection{Performance by Description Length}
\label{app:description_length}
\begin{figure*}[t]
    \centering
    \includegraphics[width=0.85\linewidth]{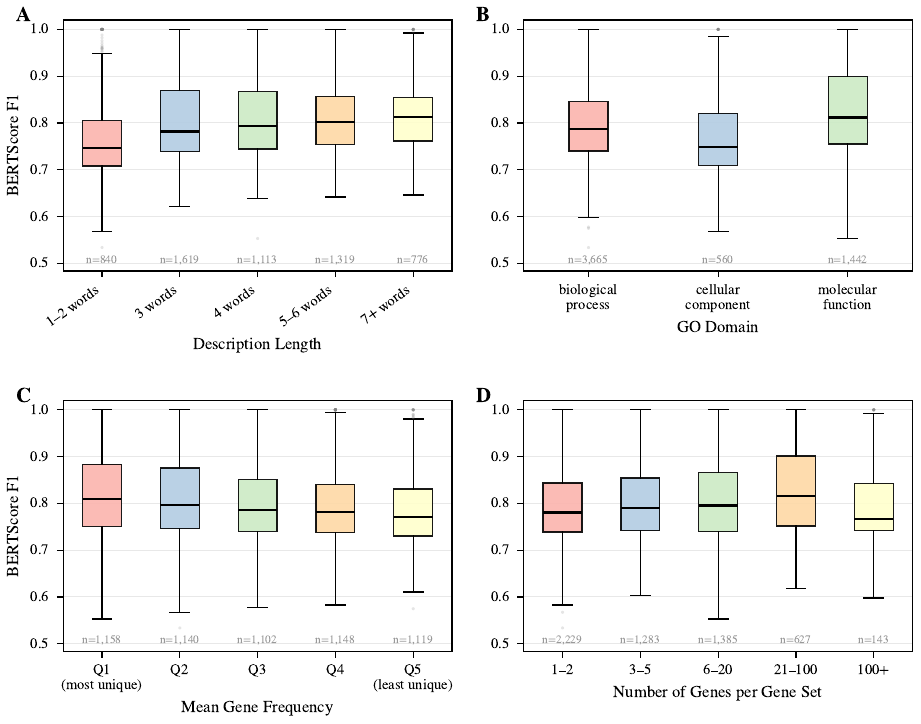}
    \vspace{-0.2cm}
    \caption{BERTScore F1 performance across gene set characteristics in the GO dataset. Box plots display the distribution of BERTScore F1 scores stratified by four gene set properties: \textbf{(A)} description length in words, \textbf{(B)} GO domain (biological process, cellular component, molecular function), \textbf{(C)} mean gene frequency across all gene sets (binned into quintiles Q1–Q5, where Q1 represents the most unique genes and Q5 the least unique), and \textbf{(D)} number of genes per gene set. Sample sizes (n) are indicated at the bottom of each panel.}
    \vspace{-0.3cm}
    \label{fig:go_box_plots}
\end{figure*}

We analyze model performance as a function of annotation length, stratified by the number of 
words in the reference description, for both the GO and MSigDB datasets in \textbf{(A)} of Figures \ref{fig:go_box_plots} and \ref{fig:msigdb_box_plots}. On GO, performance 
is relatively stable across short and mid-length descriptions (1–6 words), with a modest 
improvement for longer annotations (7+ words), suggesting that more descriptive ground-truth 
labels are easier for the model to approximate. On MSigDB, the trend is reversed: shorter 
descriptions (1–2 words) yield the highest BERTScore F1, with performance declining as 
description length increases. This may reflect the fact that short MSigDB annotations tend 
to describe well-defined, canonical pathways that are more straightforwardly predicted, 
whereas longer descriptions capture more complex or nuanced biological processes. Together, 
these results suggest that annotation complexity is a meaningful factor in model performance, 
and that the relationship between description length and difficulty varies across databases.
\subsection{Performance by Gene Set Cardinality}
\label{app:cardinality}
\begin{figure*}[t]
    \centering
    \includegraphics[width=0.9\linewidth]{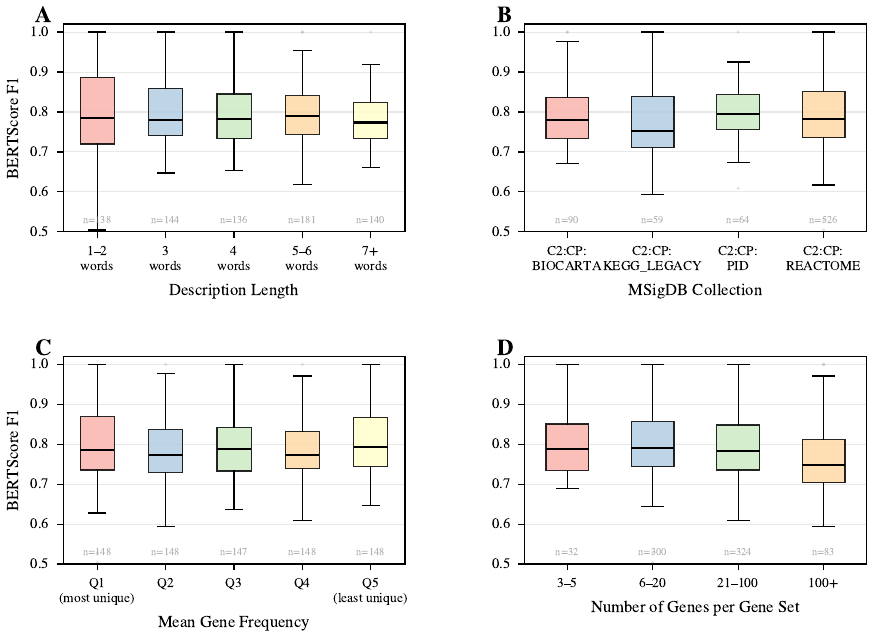}
    \caption{BERTScore F1 performance across gene set characteristics in the MSigDB dataset. Box plots display the distribution of BERTScore F1 scores stratified by four gene set properties: \textbf{(A)} description length in words, \textbf{(B)} MSigDB collection (BIOCARTA, KEGG LEGACY, PID, REACTOME), \textbf{(C)} mean gene frequency across all gene sets (binned into quintiles Q1–Q5, where Q1 represents the most unique genes and Q5 the least unique), and \textbf{(D)} number of genes per gene set. Sample sizes (n) are indicated at the bottom of each panel.}
    \label{fig:msigdb_box_plots}
\end{figure*}

We analyze model performance as a function of gene set cardinality across both datasets in \textbf{(D)} of Figure \ref{fig:go_box_plots} and \ref{fig:msigdb_box_plots}. 
On GO, performance increases monotonically with set size, with the exception of the largest gene sets 
(100+ genes) achieving noticeably lower BERTScore than smaller sets.
Larger gene sets may provide more aggregate sequence information, although differences in annotation composition and difficulty across size bins may also contribute to this trend.

On MSigDB, the trend is reversed, with performance decreasing monotonically
across cardinality bins. This may reflect the fact that 
larger MSigDB gene sets tend to describe broader, more complex 
pathway-level concepts that are inherently harder to summarize 
concisely, whereas smaller sets often correspond to well-defined, 
canonical pathways with more predictable annotations. Together, 
these results suggest that the relationship between set size and 
annotation difficulty is database-dependent, reflecting the 
structural differences between GO and MSigDB annotation schemas.
\subsection{Performance by Domain and Collection}
\label{app:domain_collection}
We examine performance across biological domains in GO and collection sources in MSigDB. The results are presented in \textbf{(B)} of Figure \ref{fig:go_box_plots} and \ref{fig:msigdb_box_plots}. 
Across GO domains, molecular function annotations achieve the highest BERTScore, followed by biological process, with cellular component yielding the lowest scores. However, absolute performance does not directly reflect the contribution of protein-sequence representations. As shown by the per-domain PLM ablation in Table~\ref{tab:domain_ablation}, the benefit of protein embeddings varies across domains, with negligible gains for biological process, modest gains for molecular function, and the largest improvements for cellular component. These results suggest that sequence-derived representations provide complementary information whose contribution depends on the biological domain.

Across MSigDB collections, the PID collection achieves the strongest performance despite its small number of examples (n=64).
BIOCARTA, KEGG LEGACY, and REACTOME show comparable and slightly lower performance, with higher variance, potentially reflecting 
greater heterogeneity in annotation style and the smaller number of examples available 
for each collection.

The performance differences observed across GO domains and MSigDB collections
are consistent with the differing compositional structure of each database.
GO terms group genes that share a specific annotated function by definition,
meaning gene membership alone is already highly predictive of the annotation
and leaves limited room for protein sequence embeddings to provide additional
signal beyond what co-occurrence patterns already encode. MSigDB gene sets
are instead drawn from heterogeneous experimental signatures across diverse
sources, with their authors explicitly noting heterogeneity within gene sets;
in this setting, symbolic gene identity is a less reliable cue and
aggregating protein sequence information across diverse gene members provides
a substantially stronger discriminative signal. This also explains the
asymmetry observed in the PLM embedding ablation (Table \ref{tab:ablations}), where protein
embeddings provide a larger performance gain on MSigDB than GO, which is a pattern that is biologically interpretable rather than anomalous. The gene-disjoint generalization results (Table \ref{tab:gene_disjoint_results}) further support this interpretation that SoftGene exploits transferable sequence-level features rather than memorizing gene identity or co-occurrence patterns.
\subsection{Performance by Gene Uniqueness}
\label{app:gene_uniqueness}
To determine if model performance is influenced by the specificity of the underlying genes, we analyzed BERTScore F1 as a function of Mean Gene Frequency, stratified into quintiles from Q1 (most unique) to Q5 (least unique).

For a gene set $\mathcal{G}_i$ with genes $g \in \mathcal{G}_i$, we define mean gene frequency as
\begin{equation}
\mathrm{MGF}(\mathcal{G}_i)=\frac{1}{\lvert\mathcal{G}_i\rvert}\sum_{g\in\mathcal{G}_i} f(g),
\end{equation}
where $f(g)$ denotes the number of gene sets in the dataset that contain gene $g$.

The results are presented in \textbf{(C)} of Figure \ref{fig:go_box_plots} and \ref{fig:msigdb_box_plots}. In GO, median performance exhibits a gradual decline from Q1 to Q5, with modestly higher variance in lower quintiles, indicating that gene sets composed of highly unique genes are somewhat more variable in difficulty. In MSigDB, median performance remains largely stable across Q1--Q4, followed by a slight increase in Q5. This pattern suggests that sets containing frequently reused (i.e., less unique) genes may be marginally easier to annotate, potentially because such genes are enriched in well-characterized canonical pathways.

Overall, the effect of gene uniqueness is limited relative to the broader performance range, indicating that SoftGene remains robust across varying levels of gene reuse.
\subsection{Performance by Ontology Depth}
\label{app:ontology_depth}
\begin{figure}[t]
    \centering
    \includegraphics[width=0.9\linewidth]{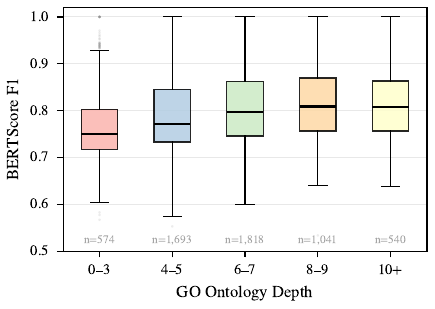}
    \caption{BERTScore F1 performance stratified by GO ontology depth. Box plots display the distribution of BERTScore F1 scores across gene sets binned by their depth within the Gene Ontology hierarchy (0–3, 4–5, 6–7, 8–9, and 10+).}
    \label{fig:go_depth_box_plot}
\end{figure}

\vspace{-3pt}
We further evaluate model performance on the GO dataset stratified by ontology depth, which serves as a proxy for functional specificity. As shown in Figure \ref{fig:go_depth_box_plot}, BERTScore tends to increase with ontology depth. This upward trend suggests that the model is more effective at approximating the descriptions of highly specific biological terms than those of broader, more generic categories located near the root of the ontology. This may be due to more specific terms having distinct, well-defined lexical profiles in the training data compared to the abstract and often overlapping definitions of high-level terms.
\begin{table*}[ht]
\centering
\caption{Performance of SoftGene and baselines on the Gene Ontology (GO) and MSigDB datasets using additional performance metrics. Best results are indicated by \textbf{boldface}.}
\label{tab:additional_metrics}
\small
\begin{tabular}{lcccccc}
\toprule
\multirow{2}{*}{\textbf{Configuration}} & \multicolumn{2}{c}{\textbf{Gene Ontology (GO)}} & \multicolumn{1}{c}{\textbf{MSigDB}} \\
\cmidrule(lr){2-4} \cmidrule(lr){5-7}
 & Semantic Similarity & Hierarchical Semantic Similarity & Semantic Similarity \\
\midrule
\multicolumn{7}{c}{\textbf{Prompt Engineering Methods}} \\
\midrule
Zero-shot    & 0.861 \tiny{± 0.002}  & 0.871 \tiny{± 0.002} & 0.876 \tiny{± 0.001} \\
Few-shot & 0.860 \tiny{± 0.001}  & 0.869 \tiny{± 0.001} & 0.872 \tiny{± 0.002} \\
One-shot CoT & 0.863 \tiny{± 0.010}  & 0.885 \tiny{± 0.000} & 0.876 \tiny{± 0.003} \\
\midrule
\multicolumn{7}{c}{\textbf{Prompt-tuning Methods}} \\
\midrule
Gene Symbol Soft Prompting & 0.874 \tiny{± 0.001}  & 0.897 \tiny{± 0.001} & 0.879 \tiny{± 0.003} \\
P-Tuning    & 0.882 \tiny{± 0.002}  & 0.896 \tiny{± 0.002} & 0.868 \tiny{± 0.003} \\
Prefix Tuning    & 0.877 \tiny{± 0.004}  & 0.890 \tiny{± 0.003} & 0.886 \tiny{± 0.001} \\
\midrule
\multicolumn{7}{c}{\textbf{Agentic Systems}} \\
\midrule
Discovera & 0.865 \tiny{± 0.001}  & 0.893 \tiny{± 0.001} & 0.874 \tiny{± 0.000} \\
TALISMAN & 0.867 \tiny{± 0.000}  & 0.892 \tiny{± 0.001} & 0.870 \tiny{± 0.004} \\
GeneAgent & 0.865 \tiny{± 0.001}  & 0.888 \tiny{± 0.001} & 0.873 \tiny{± 0.003} \\
\midrule
\multicolumn{7}{c}{\textbf{SoftGene}} \\
\midrule
SoftGene (Llama 3.1 8B) & \textbf{0.888 \tiny{± 0.003}}  & \textbf{0.900 \tiny{± 0.001}} & \textbf{0.899 \tiny{± 0.002}} \\
\bottomrule
\end{tabular}
\end{table*}
\subsection{Additional Performance Metrics}
\label{sec:additional_metrics}
To further assess biological validity beyond surface-level text similarity, we evaluate SoftGene and all other baseline methods using two biologically-grounded metrics: semantic similarity computed using PubMedBERT~\cite{gu2021domain}, which is a biomedical domain-specific encoder, and Hierarchical Semantic Similarity (HSS) (following \citet{wang2024geneagentselfverificationlanguageagent}), which measures whether predicted Gene Ontology (GO) terms are consistent with the hierarchical structure of the GO directed acyclic graph (DAG).

For HSS, we first construct a hierarchical neighborhood for each ground-truth GO term by traversing the GO DAG upward to collect ancestor terms. The traversal includes standard GO relationships such as ``is a'', ``part of'', and regulatory edges (e.g., ``regulates'', ``positively regulates'', and ``negatively regulates''), and is restricted to a maximum hop distance of $K = 3$. This yields a set of biologically related ancestor terms that represent general functional abstractions of the original annotation.

Each annotation is represented using a biomedical sentence encoder (PubMedBERT), which maps both predicted annotations $\hat{y}$ and ancestor terms $a \in \mathcal{A}_K(y)$ into a shared semantic embedding space. HSS is then defined as the maximum cosine similarity between the embedding of the predicted annotation and the embeddings of all ancestor terms:
\begin{equation}
\mathrm{HSS}(\hat{y}, y) = \max_{a \in \mathcal{A}_K(y)} \cos\left(\mathbf{e}_{\hat{y}}, \mathbf{e}_a\right),
\end{equation}
where $\mathbf{e}_{\hat{y}}$ and $\mathbf{e}_a$ denote the embeddings of the predicted annotation and ancestor term respectively. HSS is computed exclusively for the Gene Ontology dataset, as MSigDB does not define an explicit term hierarchy. Results are reported in Table~\ref{tab:additional_metrics}. SoftGene achieves the strongest performance across the biologically grounded metrics, indicating improved semantic alignment and ontology-aware consistency. However, the margins over competing methods are smaller than those observed for lexical metrics such as METEOR and ROUGE-L. This suggests that part of the larger lexical improvement may reflect better alignment with the phrasing and style of the reference annotations, while the gains in biological content are more modest. Thus, we interpret the biologically grounded metrics as complementary evidence of improved annotation quality rather than as showing improvements of the same magnitude as the lexical metrics.

\subsection{Additional Case Studies}
\label{app:case_studies}
\subsubsection{Regulation of Megakaryocyte Differentiation}
We examine the biological process gene set GO:0045652 (regulation of megakaryocyte differentiation), for which the model produces a near-perfect prediction.
The highest gene-level attention weight is assigned to MEF2C, a transcription factor known to be essential for megakaryocyte differentiation and platelet production \cite{gekas2009mef2c}.
RBM15, which directly regulates alternative splicing of key megakaryopoiesis transcription factors including GATA1 and RUNX1, receives an intermediate attention weight, consistent with its established role in this process \cite{zhang2015cross}.
In contrast, LOX and EIF6, which are a matrix-remodeling enzyme and a general translation factor, respectively, receive among the lowest attention weights, reflecting their more peripheral relevance to megakaryocyte differentiation specifically.
These results suggest that the hierarchical attention encoder successfully distinguishes between genes that are central to megakaryocyte differentiation and those with more general cellular functions within the same gene set.
\subsubsection{Hypoxia-Inducible Factor in the Cardiovascular System}
We examine the BIOCARTA\_HIF\_PATHWAY gene set, which represents hypoxia-inducible factor (HIF) signaling. The model assigns its highest attention to HIF1A, VHL, and VEGFA.
HIF1A and VHL form the central oxygen-sensing regulatory mechanism, where VHL controls degradation of HIF1A under normal oxygen conditions \cite{jaakkola2001targeting}. VEGFA is a key downstream target that drives vascular growth in response to hypoxia \cite{semenza2012hypoxia}.
In contrast, genes with more general cellular functions, such as HSP90AA1 and P4HB, receive lower attention weights, consistent with roles in protein folding and cellular maintenance rather than hypoxia-specific signaling.
Overall, the model concentrates attention on core pathway regulators and outputs, while assigning lower importance to broadly expressed housekeeping genes.
\subsubsection{Glyoxylate and Dicarboxylate Metabolism}
We examine the KEGG\_GLYOXYLATE\_ AND\_DICARBOXYLATE\_METABOLISM pathway, for which the model assigns elevated attention to several canonical metabolic enzymes. The highest attention weight is assigned to CS (citrate synthase), a core enzyme of central carbon metabolism that catalyzes entry into the tricarboxylic acid cycle \cite{zhang2025citrate}. Related enzymes including ACO1/ACO2 and MDH1/MDH2 also receive high attention weights, consistent with their established roles in interconnected dicarboxylate and mitochondrial metabolic processes \cite{omini2024tca}.
The model additionally assigns elevated attention to HAO1, HAO2, and GRHPR, which are more directly associated with glyoxylate metabolism and detoxification pathways \cite{wanders2025human, garrelfs2024glyoxylate}. Together, these patterns suggest that the model upweights metabolically related enzymes within the pathway rather than assigning attention uniformly across all genes.
\section{Qualitative Analysis}
\label{app:qualitative_analysis}
\subsection{High- and Low-Scoring Completions}
\label{app:high_low_completions}
Tables \ref{tab:top_bottom_predictions_go} and \ref{tab:top_bottom_predictions} present the top and bottom performing gene set predictions for GO and MSigDB, respectively, as measured by BERTScore F1. High-scoring examples demonstrate the model's ability to precisely capture enzymatic activities and metabolic pathways, whereas lower-scoring cases often involve more complex regulatory processes or nuanced phenotypic descriptions.
\begin{table*}[t]
\centering
\caption{Top 5 and bottom 5 scoring gene set predictions from GO by BERTScore F1 (excluding perfect scores).}
\label{tab:top_bottom_predictions_go}
\scriptsize
\setlength{\tabcolsep}{3pt}
\renewcommand{\arraystretch}{1.1}
\begin{tabular}{lp{4.2cm}cp{3.8cm}p{3.8cm}}
\toprule
\textbf{Group} & \textbf{Gene Set} & \textbf{Score} & \textbf{Reference} & \textbf{Predicted} \\
\midrule
\multirow[c]{5}{*}{\rotatebox[origin=c]{90}{\textbf{Top 5}}}
& GO:0004143 & 0.9500 & ATP-dependent diacylglycerol kinase activity & diacylglycerol kinase activity \\
& GO:0045588 & 0.9500 & positive regulation of gamma-delta T cell differentiation & positive regulation of T cell differentiation \\
& GO:0042129 & 0.9499 & regulation of T cell proliferation & positive regulation of T cell activation \\
& GO:0140950 & 0.9496 & histone H2A deubiquitinase activity & histone H2A deubiquitination \\
& GO:0006563 & 0.9493 & L-serine metabolic process & serine metabolic process \\
\midrule
\multirow[c]{5}{*}{\rotatebox[origin=c]{90}{\textbf{Bottom 5}}}
& GO:0030431 & 0.5774 & sleep & positive regulation of interleukin-18 production \\
& GO:0043276 & 0.5747 & anoikis & positive regulation of cell cycle G1/S phase transition \\
& GO:0014705 & 0.5672 & C zone & myosin binding \\
& GO:0030628 & 0.5530 & pre-mRNA 3'-splice site binding & U2 snRNP \\
& GO:1990603 & 0.5337 & dark adaptation & regulation of G protein-coupled receptor signaling pathway \\
\bottomrule
\end{tabular}
\vspace{0.3cm}
\end{table*}
\begin{table*}[t]
\centering
\caption{Top 5 and bottom 5 scoring gene set predictions from MSigDB by BERTScore F1 (excluding perfect scores).}
\label{tab:top_bottom_predictions}
\scriptsize
\setlength{\tabcolsep}{3pt}
\renewcommand{\arraystretch}{1.1}
\begin{tabular}{lp{4.2cm}cp{3.8cm}p{3.8cm}}
\toprule
\textbf{Group} & \textbf{Gene Set} & \textbf{Score} & \textbf{Reference} & \textbf{Predicted} \\
\midrule
\multirow[c]{5}{*}{\rotatebox[origin=c]{90}{\textbf{Top 5}}}
& \shortstack[l]{REACTOME\_SARS\_COV\_INFECTIONS} & 0.9477 & SARS-CoV Infections & SARS-CoV-2 infection \\
& \shortstack[l]{REACTOME\_DOWNSTREAM\_\\SIGNALING\_OF\_ACTIVATED\_\\FGFR4} & 0.9458 & Downstream signaling of activated FGFR4 & Signaling by FGFR4 \\
& \shortstack[l]{REACTOME\_DOWNSTREAM\_\\SIGNALING\_OF\_ACTIVATED\_\\FGFR2} & 0.9454 & Downstream signaling of activated FGFR2              & Signaling by FGFR2 \\
& \shortstack[l]{BIOCARTA\_MTOR\_PATHWAY} & 0.9453 & mTOR Signaling Pathway  & mTOR signaling \\
& \shortstack[l]{REACTOME\_METHIONINE\_SALVAGE\_\\PATHWAY} & 0.9450 & Methionine salvage pathway & Methionine salvage \\
\midrule
\multirow[c]{5}{*}{\rotatebox[origin=c]{90}{\textbf{Bottom 5}}}
& \shortstack[l]{REACTOME\_S\_PHASE} & 0.6173 & S Phase & Regulation of APC/C activators in early mitosis \\
& \shortstack[l]{PID\_ALPHA\_SYNUCLEIN\_PATHWAY} & 0.6088 & Alpha-synuclein signaling & Parkinson's disease \\
& \shortstack[l]{KEGG\_HUNTINGTONS\_DISEASE} & 0.5938 & Huntington's disease & Mitochondrial ATP synthase complex \\
& \shortstack[l]{REACTOME\_MISCELLANEOUS\_\\SUBSTRATES} & 0.5279 & Miscellaneous substrates & CYP4A11, CYP4A22, CYP4B1, ... \\
& \shortstack[l]{REACTOME\_FATTY\_ACIDS} & 0.5047 & Fatty acids & CYP4A11, CYP4A22, \ldots \\
\bottomrule
\end{tabular}
\end{table*}
\subsection{Failure Mode Analysis}
\label{app:failure_analysis}
While SoftGene achieves strong performance across both benchmarks, 
a qualitative analysis of low-scoring predictions reveals systematic 
and interpretable failure patterns. Rather than reflecting random 
errors, these cases illuminate specific limitations of the model's 
learned representations and, in some instances, expose shortcomings 
of the evaluation metrics themselves. We identify three recurring 
failure modes below.
\subsubsection*{Substrate Specificity Failure}
\noindent\textbf{Example Gene Set:} GO:0015756 \\
\noindent\textbf{Predicted:} "sodium:sugar symporter" \\
\noindent\textbf{Reference:} "fucose transmembrane transport" \\
\noindent\textbf{Analysis}: This failure mode involves the accurate identification of a molecular mechanism while failing to specify the precise chemical substrate. This example suggests that while the PLM embeddings may encode structural signals associated with the transporter class, the gene set encoder may aggregate residue-level signals to an extent that subtle active-site configurations are smoothed over. Consequently, the model prioritizes robust functional class recognition over minute chemical variations.
\subsubsection*{Pathway Pleiotropy Failure}
\noindent\textbf{Example Gene Set}: GO:0045743 \\
\noindent\textbf{Predicted}: "cell-substrate adhesion"\\
\noindent\textbf{Reference}: "fibroblast growth factor receptor signaling"\\
\noindent\textbf{Analysis}: The model occasionally identifies biological processes that are functionally coupled with, or downstream of, the primary reference term. In the case of the example, biologically, fibroblast growth factor receptors are primary regulators of focal adhesions and cytoskeletal remodeling \cite{medsci1010020,eswarakumar2005cellular}. These results indicate that the soft prompt tokens are capturing a broader, pleiotropic signature of the gene set. This suggests the model is identifying a valid biological "neighborhood" or a high-level structural consequence of the signaling pathway, suggesting that the learned representation may capture broader functional relationships among cellular processes.
\subsubsection*{Semantic Granularity Failure}
\noindent\textbf{Example Gene Set}: GO:0045187 \\
\noindent\textbf{Predicted}: "regulation of circadian rhythm" \\
\noindent\textbf{Reference}: "regulation of circadian sleep/wake cycle" \\
\noindent\textbf{Analysis}: This failure mode highlights instances where the model generates semantically valid synonyms or parent terms that are penalized by rigid string-matching metrics. While the predicted and reference terms are functionally nearly identical, the resulting ROUGE-L score is low despite a high BERTScore. This discrepancy underscores a limitation of standard NLP evaluation in the biological domain and confirms that SoftGene is capturing the core biological concept with high fidelity, even when the generated output diverges from the specific granularity of the curated ground truth.
\section{Implementation Details}
\label{app:implementation_details}
We use the Hugging Face Transformers library \cite{wolf2019huggingface} for all local LLM inference and training, the Accelerate library \cite{accelerate} for distributed training, and the Evaluate library \cite{von-werra-etal-2022-evaluate} for evaluation metrics. For the BERTScore metric, we use DistilBERT \cite{sanh2019distilbert}. We utilize a server equipped with three NVIDIA A6000 GPUs, 512 GB of RAM, and a 48-core CPU.
\subsection{Hyperparameters}
\label{sec:hyperparameters}
The hyperparameters used for training SoftGene on the GO and MSigDB datasets are presented in Table~\ref{tab:hyperparameters}.
To obtain $\boldsymbol{H}_{cont}$, LLM inference was conducted using the default generation hyperparameters, with the sole modification that the number of max new tokens was set to 100.
\subsection{Baselines}
\label{app:baselines}
\subsubsection*{Prompt Engineering Methods}

For all prompt engineering baselines, we use the \texttt{Llama-3.1-8B-Instruct} model. We conducted inference via the Replicate API using the default inference configuration.

\subsubsection*{Prompt-Tuning Methods}

For all prompt-tuning baselines, we use the base \texttt{Llama-3.1-8B} model. We employ soft prompt tuning with 10 virtual tokens. We omit commercial LLM context ($H_{cont}$) for this setting, but note that SoftGene outperforms prompt tuning baselines even without this context (see "w/o Context ($H_{cont}$)" in Table \ref{tab:ablations}). Models were trained for one epoch on the GO dataset and two epochs on the MSigDB dataset. Due to the short length of both the prompts and target gene set annotations, prompt tuning converged rapidly. 

\subsubsection*{Agentic Systems}

For all agentic system baselines, we use the \texttt{GPT-4o-mini} model accessed via the OpenAI API. To ensure comparable inference costs with SoftGene, we limit the number of self-verification iterations in the GeneAgent baseline to three. Similarly, for Discovera, we restrict the number of genes passed to the INDRA API to 10, as the API returns pairwise interactions between genes, which otherwise leads to substantially longer inference times.

\subsection{Runtime Analysis}
\label{app:runtime_analysis}
In this section, we report the runtime for each phase of the SoftGene framework. Protein 
embedding inference is performed once over the entire human proteome prior to training and 
cached for all subsequent use; runtime details for this step are reported in Table \ref{tab:runtime_plm}. 
Because protein embeddings are precomputed and cached, no repeated PLM forward pass is required during SoftGene training or evaluation; 
the runtime cost of the hierarchical attention encoder is limited to a linear-time sparse scatter 
aggregation and a small MLP projection, both of which are negligible relative to the LLM forward 
pass. Training time and memory details are reported in Table~\ref{tab:runtime_train}. Evaluation 
time and memory details are reported in Table~\ref{tab:runtime_eval}. We note that for novel 
gene sets not present in GO or MSigDB, an additional commercial LLM inference pass is required to 
generate the auxiliary context $\boldsymbol{H}_{cont}$; we omit this from our runtime 
reporting due to its highly variable network latency.

\begin{table}[ht]
\centering
\small
\caption{Protein embedding inference runtime using ESM-2 on a single NVIDIA A6000 GPU. \label{tab:runtime_plm}}
\begin{tabular}{ll}
\toprule

\textbf{Metric} & \textbf{Value} \\
\midrule
Total Time & 6{,}575.92s (1.83 hr) \\
Average Time per Protein & 26.80 ms \\
Throughput & 37.3 proteins/s \\
Peak GPU Memory & 2.250 GB \\
\bottomrule
\end{tabular}

\end{table}

\begin{table}[ht]
\centering
\small
\caption{SoftGene training runtime on the GO dataset using two NVIDIA A6000 GPUs. \label{tab:runtime_train}}
\begin{tabular}{ll}
\toprule
\textbf{Metric} & \textbf{Value} \\
\midrule
Total Training Time & 1{,}709.11s (0.47 hr) \\
Peak GPU Memory & 46.39 GB \\
\bottomrule
\end{tabular}

\end{table}

\begin{table}[ht]
\small
\caption{Inference runtime comparison between SoftGene and the zero-shot baseline on the GO 
dataset using a single NVIDIA A6000 GPU. \label{tab:runtime_eval}}
\resizebox{\columnwidth}{!}{%
\centering
\begin{tabular}{lcc}
\toprule
\textbf{Metric} & \textbf{SoftGene} & \textbf{Zero-Shot} \\
\midrule
Total Inference Time & 584.72s & 478.91s \\
Average Time per Sample & 103.18 ms & 84.51 ms \\
Throughput & 9.69 samples/s & 11.83 samples/s \\
Peak GPU Memory & 19.84 GB & 18.87 GB \\
\bottomrule
\end{tabular}}

\end{table}
\begin{table}[ht]
\centering
\caption{Inference runtime of SoftGene on the five largest gene sets in the GO evaluation set.\label{tab:runtime_large}}
\small
\begin{tabular}{ll}
\toprule
\textbf{Metric} & \textbf{Value} \\
\midrule
Total Time & 2.43 s \\
Average Time per Sample & 467.45 ms \\
Throughput & 2.14 samples/s \\
\bottomrule
\end{tabular}

\end{table}
\subsubsection{Scalability to Large Gene Sets}
A potential concern is the scalability of SoftGene to large gene sets, as the GO dataset 
contains gene sets with up to 13,616 member genes (Table~\ref{tab:statistics}). Our 
attention pooling operates in $\mathcal{O}(n)$ time with respect to gene set size, as we 
employ query-only attention rather than pairwise attention, avoiding the quadratic memory 
bottleneck common to standard self-attention. We additionally leverage sparse scatter 
operations to perform efficient aggregation across variable-length gene sets in a single 
batched operation. To empirically validate this, we report inference runtime on the five 
largest gene sets in our evaluation set in Table~\ref{tab:runtime_large}, observing that 
even at this scale SoftGene remains computationally tractable in this setting.

\section{Dataset Preparation}
\label{app:dataset_preparation}
In this section, we outline the steps taken to prepare the  benchmark datasets.
\subsection{Protein Sequences and Isoforms}
\label{app:protein_sequences}
We obtain protein sequences from the Ensembl proteome (\texttt{Homo\_sapiens.GRCh38.pep.all.fa}\footnote{\url{https://useast.ensembl.org/info/data/ftp/index.html}}) \cite{martin2023ensembl}. This resource also provides a mapping from gene symbols to their corresponding splice isoforms, giving us the hierarchical gene-to-protein information required for the gene set annotation task. The Ensembl proteome thus provides all necessary information to construct input representations for both the GO and MSigDB datasets, enabling the model to leverage gene- and protein-level features consistently across benchmarks.
\subsection{GO Dataset}
\label{app:go_dataset}
To construct the GO gene sets, we use two primary input files: the GO ontology file (\texttt{go.obo}\footnote{\url{https://geneontology.org/docs/download-ontology/}}) and the human gene annotation file (\texttt{goa\_human.gaf}\footnote{\url{https://current.geneontology.org/products/pages/downloads.html}}) obtained from the GO Consortium. The ontology file defines all GO terms and their hierarchical relationships, while the GAF file provides gene-to-term annotations. 

We load the ontology using \texttt{GODag} from the \texttt{goatools} library and read the GAF following the GAF v2.2 format. For each GO term, we collect all associated human gene symbols, storing the term's name and namespace (biological process, molecular function, or cellular component). Finally, gene sets are exported as a JSON object mapping each GO ID to its genes, description, and domain.
\subsection{MSigDB Dataset}
\label{app:msigdb_dataset}
To construct the MSigDB gene sets from the C2:CP collection, we use two input sources: the GMT file (\texttt{c2.all.v2025.1.Hs.symbols.gmt}\footnote{\url{https://www.gsea-msigdb.org/gsea/msigdb/collections.jsp}}) and the accompanying SQLite database (\texttt{msigdb\_v2025.1.Hs.db}\footnote{\url{https://www.gsea-msigdb.org/gsea/downloads.jsp}}). The GMT file provides the gene sets as lists of gene symbols, while the database contains metadata including gene set descriptions and collection information.

We first read the GMT file, extracting gene set names and their associated gene symbols. Next, we query the database to retrieve brief descriptions and collection names for all C2:CP gene sets, excluding certain sub-collections (e.g., WIKIPATHWAYS and KEGG\_MEDICUS) to focus on canonical pathway sets. Gene sets are filtered to retain only those present in both the GMT file and the database. The gene sets, including cleaned descriptions, gene symbols, and collection names, are exported as a structured JSON object mapping gene set names to their content.
\subsection{Dataset Contamination Analysis}
\label{app:contamination}
Data contamination is a common concern in NLP and LLM research, particularly when evaluation 
datasets are drawn from publicly available resources that may overlap with model pretraining 
corpora. We address two potential sources of contamination relevant to our evaluation.

\paragraph{Memorization of GO and MSigDB annotations.} We construct the datasets from late-2025 releases of GO and MSigDB. For GPT-4.1 nano, the documented knowledge cutoff predates these releases, reducing the possibility that release-specific updates were included in its pretraining data. However, many underlying genes, pathways, and annotations predate the selected releases and may have appeared in model pretraining corpora. Therefore, the use of recent database releases reduces but does not eliminate the possibility of contamination.

\paragraph{Data leakage through auxiliary context.} A separate concern is whether 
$\boldsymbol{H}_{cont}$ encodes ground-truth label information, as commercial LLMs may have 
memorized annotations and pathway descriptions from biomedical literature. We address this 
in two ways. First, as shown in Table~\ref{tab:ablations}, SoftGene without auxiliary context 
(``w/o Context $\boldsymbol{H}_{cont}$'') still outperforms all competitive baselines across 
both datasets. If $\boldsymbol{H}_{cont}$ were truly encoding ground-truth label information, removing 
it would cause a catastrophic performance drop; instead, we observe only a moderate 
decrease, confirming that it provides supplementary biological context rather than label 
information. Second, we directly measured lexical overlap between GPT-generated context and 
ground-truth annotations across both datasets, reported in Table~\ref{tab:contamination}. The 
negligible overlap confirms that observed performance cannot be attributed to verbatim copying 
from the generated context.

\begin{table}[ht]
\centering
\caption{Lexical overlap between GPT-generated auxiliary context and ground-truth annotations.}
\label{tab:contamination}
\small
\begin{tabular}{lc}
\toprule
\textbf{Dataset} & \textbf{Ground Truth Overlap} \\
\midrule
GO      & 1.37\% \\
MSigDB  & 3.41\% \\
\bottomrule
\end{tabular}
\vspace{-0.5cm}
\end{table}
\begin{table}[t]
  \centering
  \caption{Hyperparameter settings.}
  \label{tab:hyperparameters}
  \small
  \begin{tabular}{lcc}
    \toprule
    \textbf{Parameter} & \textbf{GO} & \textbf{MSigDB} \\
    \midrule
\multicolumn{3}{c}{\textbf{Hierarchical Attention Encoder}} \\
\midrule
    Learning Rate $\alpha$ & $1 \times 10^{-3}$ & $1 \times 10^{-3}$ \\
    Protein Pooling Hidden Size & 512 & 512 \\
    Gene Pooling Hidden Size & 512 & 512 \\
    Weight Decay & $1 \times 10^{-4}$ & $1 \times 10^{-4}$ \\
    Gradient Clipping Norm & 1 & 1 \\
    \midrule
\multicolumn{3}{c}{\textbf{Hybrid Soft Prompting Module}} \\
\midrule
    Training Epochs & 1 & 5 \\
    Batch Size & 4 & 4 \\
    Gradient Accumulation Steps & 4 & 4 \\
    Learning Rate $\beta$ & $5 \times 10^{-4}$ & $5 \times 10^{-4}$ \\
    Hidden Dimension & $4096$ & $4096$ \\
    Weight Decay & $1 \times 10^{-2}$ & $1 \times 10^{-2}$ \\
    Dropout Rate & $1 \times 10^{-1}$ & $1 \times 10^{-1}$ \\
    Soft Prompt Length & 10 & 10 \\
    \bottomrule
  \end{tabular}
\end{table}
\section{LLM Prompts}
\label{app:llm_prompts}
In this section, we share the LLM prompts used in our method as well as the prompts used for each of the prompting baselines.
\subsection{SoftGene Prompt}
\label{app:softgene_prompt}
Figure~\ref{fig:softgene_prompt} shows the prompt used for SoftGene. We deliberately keep the prompt concise, providing minimal details, so that the model can freely learn task-relevant information during training. This approach allows the soft prompt to adapt effectively to the gene prediction task.
\subsection{Zero-Shot Prompt}
\label{app:zeroshot_prompt}
Figure~\ref{fig:zero_shot_prompt} shows the zero-shot prompt. In this setup, we provide only task instructions, forcing the model to rely entirely on its pretraining knowledge to perform the task. No examples or task-specific demonstrations are given, making this a baseline for evaluating the model's generalization abilities.
\subsection{Few-Shot Prompt}
\label{app:fewshot_prompt}
Figure~\ref{fig:few_shot_prompt} shows the few-shot prompt. In this setup, we provide the model with a small number of task examples along with the instructions, allowing it to infer the desired output format and style. This few-shot context helps the model leverage both its pretraining knowledge and the provided demonstrations to perform the task more accurately.
\subsection{One-Shot Chain-of-Thought Prompt}
\label{app:cot_prompt}
Figure~\ref{fig:cot_prompt} shows the one-shot chain-of-thought prompt. This prompt includes a single example that demonstrates step-by-step reasoning, guiding the model to analyze the system before producing a final output. By providing both the instructions and a worked example, the prompt encourages the model to follow a reasoning process.
\begin{figure*}[p]
    \centering
    \captionsetup{font=small, skip=3pt}

    \begin{minipage}{\textwidth}
        \centering
        \begin{tcolorbox}[
            width=\textwidth,
            title=Instruction Prompt,
            colback=gray!10,
            colframe=gray!100,
            fonttitle=\bfseries\footnotesize,
            fontupper=\footnotesize,
            boxsep=0pt,
            left=3pt,
            right=3pt,
            top=1pt,
            bottom=1pt,
            before skip=1pt,
            after skip=1pt
        ]
        Input:\\
        Genes: \{genes\}\\\\
        \{commercial\_llm\_context\}\\\\
        Task: Predict GO term.\\\\
        Output:
        \end{tcolorbox}

        \captionof{figure}{Instruction prompt used for SoftGene.}
        \label{fig:softgene_prompt}
    \end{minipage}

    \vspace{3pt}

    \begin{minipage}{\textwidth}
        \centering

        \begin{tcolorbox}[
            width=\textwidth,
            title=System Prompt,
            colback=teal!10,
            colframe=teal!70,
            fonttitle=\bfseries\footnotesize,
            fontupper=\footnotesize,
            boxsep=0pt,
            left=3pt,
            right=3pt,
            top=1pt,
            bottom=1pt,
            before skip=1pt,
            after skip=2pt
        ]
        You are an efficient and insightful assistant to a molecular biologist.
        \end{tcolorbox}

        \begin{tcolorbox}[
            width=\textwidth,
            title=Human Prompt,
            colback=red!10,
            colframe=red!50,
            fonttitle=\bfseries\footnotesize,
            fontupper=\footnotesize,
            boxsep=0pt,
            left=3pt,
            right=3pt,
            top=1pt,
            bottom=1pt,
            before skip=0pt,
            after skip=1pt
        ]
        You are a knowledgeable bioinformatics assistant. Given a set of related genes, generate one concise biological pathway or
molecular function label that represents the overall shared function of the entire gene set, not individual genes.\\\\
The output must:\\
- Be a single short description (3–5 words)\\
- Capture the common biological theme\\
- Contain only the description, no explanation or extra text\\\\
Now complete the task:\\
\{gene\_symbols\}
        \end{tcolorbox}

        \captionof{figure}{System and user prompt used for zero-shot prompting baseline.}
        \label{fig:zero_shot_prompt}
    \end{minipage}

    \vspace{3pt}

    \begin{minipage}{\textwidth}
        \centering

        \begin{tcolorbox}[
            width=\textwidth,
            title=System Prompt,
            colback=teal!10,
            colframe=teal!70,
            fonttitle=\bfseries\footnotesize,
            fontupper=\footnotesize,
            boxsep=0pt,
            left=3pt,
            right=3pt,
            top=1pt,
            bottom=1pt,
            before skip=1pt,
            after skip=2pt
        ]
        You are an efficient and insightful assistant to a molecular biologist.
        \end{tcolorbox}

        \begin{tcolorbox}[
            width=\textwidth,
            title=Human Prompt,
            colback=red!10,
            colframe=red!50,
            fonttitle=\bfseries\footnotesize,
            fontupper=\footnotesize,
            boxsep=0pt,
            left=3pt,
            right=3pt,
            top=1pt,
            bottom=1pt,
            before skip=0pt,
            after skip=1pt
        ]
        You are a knowledgeable bioinformatics assistant. Given a set of related genes, generate one concise biological pathway or
        molecular function label that represents the overall shared function of the entire gene set, not individual genes.\\\\
        The output must:\\
        - Be a single short description (3–5 words)\\
        - Capture the common biological theme\\
        - Contain only the description, no explanation or extra text\\\\
        Consider the following examples:\\
        \{examples\}\\\\
        Now complete the task:\\
        \{gene\_symbols\}
        \end{tcolorbox}

        \captionof{figure}{System and user prompt used for few-shot prompting baseline.}
        \label{fig:few_shot_prompt}
    \end{minipage}

    \vspace{3pt}

    \begin{minipage}{\textwidth}
        \centering

        \begin{tcolorbox}[
            width=\textwidth,
            title=System Prompt,
            colback=teal!10,
            colframe=teal!70,
            fonttitle=\bfseries\footnotesize,
            fontupper=\footnotesize,
            boxsep=0pt,
            left=3pt,
            right=3pt,
            top=1pt,
            bottom=1pt,
            before skip=1pt,
            after skip=2pt
        ]
        You are an efficient and insightful assistant to a molecular biologist.
        \end{tcolorbox}

        \begin{tcolorbox}[
            width=\textwidth,
            title=Human Prompt,
            colback=red!10,
            colframe=red!50,
            fonttitle=\bfseries\footnotesize,
            fontupper=\footnotesize,
            boxsep=0pt,
            left=3pt,
            right=3pt,
            top=1pt,
            bottom=1pt,
            before skip=0pt,
            after skip=1pt
        ]
        Write a critical analysis of the biological processes performed by this system of interacting genes. Base your analysis on prior
        knowledge available in your training data. After completing your analysis, propose a short descriptive name for the most
        prominent biological process performed by the system. Put your chosen name at the top of the analysis as 'Process: <name>’. Be
        concise, do not use unnecessary words. Be factual, do not editorialize. Be specific, avoid overly general statements such as "the
        genes are involved in various cellular processes". Avoid listing facts about individual proteins. Instead, try to group proteins with
        similar functions and discuss their interplay, synergistic or antagonistic effects, and functional integration within the system.
        Also, avoid choosing generic process names such as "Cellular Signaling and Regulation". If you cannot identify a prominent
        biological process for the proteins in the system, I want you to communicate this in your analysis and name the process: "System
        of unrelated genes"\\
        To help you in your work, I am providing an example system of interacting genes and the corresponding example analysis output.\\\\
        \{example\}\\\\
        Here are the interacting genes:\\
        Genes: \{gene\_symbols\}\\\\
        After completing your analysis, output the most relevant Gene Ontology biological process description for these proteins.
        \end{tcolorbox}

        \captionof{figure}{System and user prompt used for one-shot chain-of-thought prompting baseline.}
        \label{fig:cot_prompt}
    \end{minipage}

\end{figure*}
\section{Additional Related Works}
\label{app:related_works}
\subsection{Protein Language Models}
\label{app:protein_language_models}
Protein language models (PLMs) apply the self-supervised pretraining paradigm from NLP to amino acid sequences, learning representations of protein structure and function directly from evolutionary data.
\citet{rivesa2021biological} established the foundation for this line of work by training ESM-1b on 250 million protein sequences, demonstrating that structural and functional properties emerge from sequence-level pretraining alone without explicit structural supervision.
\citet{lin2023evolutionary} scaled this approach to 15 billion parameters with ESM-2 and ESMFold, enabling direct inference of atomic-level protein structure from primary sequence.
\citet{elnaggar2021prottrans} introduced ProtTrans, a family of transformer models pretrained on protein sequences, demonstrating that NLP pretraining strategies transfer effectively to amino acid data; ProtBERT serves as one of the encoder backbones in our ablation study.
\citet{meier2021language} showed that PLMs support zero-shot prediction of mutational effects without any task-specific supervision, demonstrating that PLM representations encode functional information beyond structural topology and motivating our use of these embeddings to distinguish functionally distinct isoforms.
\citet{rao2021msa} introduced the MSA Transformer, which interleaves row and column attention across multiple sequence alignments to incorporate co-evolutionary context, achieving substantially improved contact prediction over single-sequence models.
\citet{hayes2025simulating} presented ESM3, a multimodal generative PLM that jointly reasons over protein sequence, structure, and function, representing the frontier of PLM capability and pointing toward richer biological representations for gene annotation tasks.
\subsection{Traditional Gene Set Annotation Methods}
\label{app:traditional_gene_set}
The analysis of gene sets has a long history in functional genomics, with early methods framing the problem as statistical over-representation testing.
\citet{beissbarth2004gostat} introduced one of the first automated approaches, identifying statistically overrepresented GO terms within a gene list by comparing observed annotation frequencies against a background distribution.
While conceptually straightforward, such over-representation methods treat gene sets as unordered collections and discard magnitude and directionality information present in expression data.
\citet{subramanian2005gene} addressed this by operating on ranked gene lists derived from differential expression, using a weighted Kolmogorov-Smirnov statistic to detect coordinated shifts in gene set activity.
However, GSEA produces enrichment scores at the cohort level, limiting applicability to single-sample settings.
\citet{barbie2009systematic} extended this to individual samples by computing per-sample enrichment scores, while \citet{hanzelmann2013gsva} further generalized the approach as a non-parametric kernel estimation problem applicable across microarray and RNA-Seq platforms.
A separate line of work recognized that gene membership alone is insufficient to capture signaling complexity.
\citet{tarca2009novel} incorporated pathway topology by weighting genes according to their connectivity within curated signaling graphs, though such methods remain dependent on the completeness of curated databases \citep{liberzon2015molecular}.
Collectively, these methods share a fundamental limitation: they map gene sets to predefined categorical labels rather than generating descriptive natural language annotations, and the methods discussed above do not directly incorporate protein-sequence representations.
SoftGene is designed to address both of these gaps.